\documentclass[10pt,twocolumn,letterpaper]{article}

\usepackage[pagenumbers]{cvpr} 

\usepackage{graphicx}
\usepackage{amsmath}
\usepackage{amssymb}
\usepackage{booktabs}
\usepackage{amsfonts}       
\usepackage{url}            
\usepackage{microtype}      
\usepackage{xcolor}         
\usepackage{enumitem}
\usepackage{xspace}
\usepackage{bbm}
\usepackage{algorithm}
\usepackage{algorithmic}
\usepackage{threeparttable}
\usepackage{multirow}
\usepackage{makecell}
\usepackage{float}
\newcommand{\tabincell}[2]{\begin{tabular}{@{}#1@{}}#2\end{tabular}}

\usepackage[pagebackref,breaklinks,colorlinks]{hyperref}

\usepackage[capitalize]{cleveref}
\crefname{section}{Sec.}{Secs.}
\Crefname{section}{Section}{Sections}
\Crefname{table}{Table}{Tables}
\crefname{table}{Tab.}{Tabs.}

\def\jing{\textcolor{black}}

\def\ice{\textcolor{black}}

\definecolor{mypink}{rgb}{0.858, 0.188, 0.478}

\definecolor{qdmy}{rgb}{0,0.6,0.6}
\def\qdmy{\textcolor{black}}

\def\mD{{\mathcal D}}

\def\mF{{\mathcal F}}
\def\mG{{\mathcal G}}
\def\mH{{\mathcal H}}

\def\mL{{\mathcal L}}

\def\mX{{\mathcal X}}
\def\mY{{\mathcal Y}}

\def\0{{\bf 0}}
\def\1{{\bf 1}}

\def\bW{{\bf W}}

\def\bm{{\bf m}}

\def\bo{{\bf o}}

\def\bv{{\bf v}}
\def\bw{{\bf w}}
\def\bx{{\bf x}}

\def\bz{{\bf z}}

\def\citep{\cite}
\def\citet{\cite}

\begin{document}

\newcommand{\methodfullname}{Elastic Architecture Search\xspace}
\newcommand{\methodshortname}{EAS\xspace}

\title{Rapid Elastic Architecture Search under Specialized Classes and Resource Constraints}

\author{Jing Liu$^1$,\quad
Bohan Zhuang$^1$\footnotemark[2],\quad
Mingkui Tan$^2$\footnotemark[2],\quad
Xu Liu$^2$,\\
Dinh Phung$^1$,\quad
Yuanqing Li$^2$,\quad
Jianfei Cai$^1$\\[0.2cm]
$^1$Monash University\quad
$^2$South China University of Technology\\
}
\maketitle

\renewcommand{\thefootnote}{\fnsymbol{footnote}}
\footnotetext[2]{Corresponding author.}

\begin{abstract}
In many real-world applications, we often need to handle various deployment scenarios, where the resource constraint and the superclass of interest corresponding to a group of classes are dynamically specified. How to efficiently deploy deep models for diverse deployment scenarios is a new challenge. Previous NAS approaches seek to design architectures for all classes simultaneously, which may not be optimal for some individual superclasses. A straightforward solution is to search an architecture from scratch for each deployment scenario, which however is computation-intensive and impractical. To address this, we present a novel and general framework, called \methodfullname (\methodshortname), permitting instant specializations at runtime for diverse superclasses with various resource constraints. To this end, we first propose to effectively train an over-parameterized network via a superclass dropout strategy during training. In this way, the resulting model is robust to the subsequent superclasses dropping at inference time. Based on the well-trained over-parameterized network, we then propose an efficient architecture generator to obtain promising architectures within a single forward pass. Experiments on three image classification datasets show that \methodshortname is able to find more compact networks with better performance while remarkably being orders of magnitude faster than state-of-the-art NAS methods, \jing{
\eg, outperforming
OFA (once-for-all) by 1.3\% on Top-1 accuracy at a budget around 361M \#MAdds on ImageNet-10. More critically, \methodshortname is able to find compact architectures within 0.1 second for 50 deployment scenarios.}
\end{abstract}

\section{Introduction}
\begin{figure}[t]
    \centering
    \includegraphics[width=0.98\linewidth]{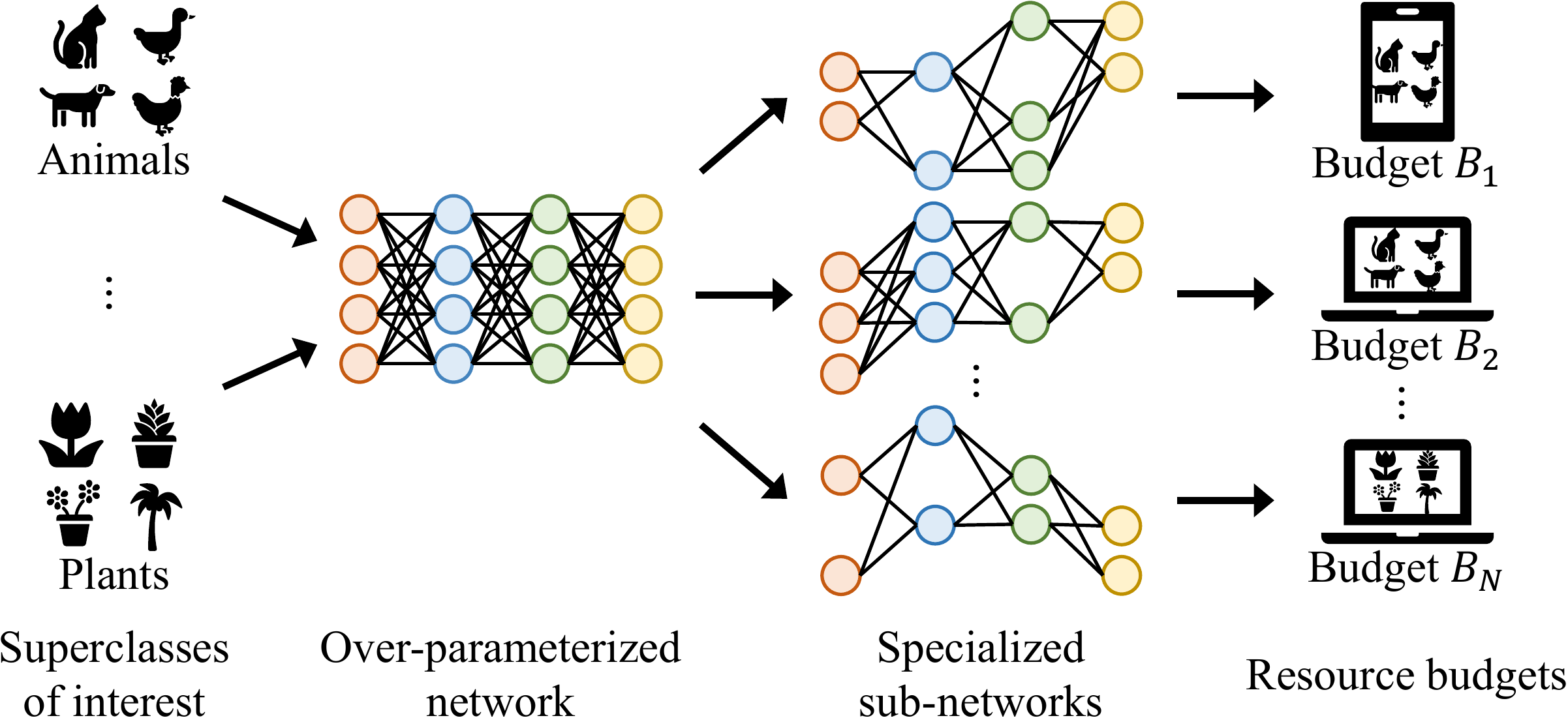}
    \caption{
    An illustration of model deployment for specified application scenarios, \eg, animal classification under various resource budgets. The optimal architectures shall be different for different deployment scenarios
    }
    \label{fig:task_overview}
\end{figure}

Deep neural networks (DNNs) have achieved state-of-the-art performance in many areas in machine learning~\cite{he2016deep,ren2015faster} and much broader in artificial intelligence. 
There are numerous practical applications where we need a well-trained general model (capable of classifying thousands of classes) to be flexible during deployment (\eg, focusing on classifying tens of classes),
as shown in Figure~\ref{fig:task_overview}. For example, zoologists might only be
interested in classifying animals such as cat, dog,
and so on, while many other classes provided by the general model might be useless to their tasks of interest. A usual additional requirement in practice is to deploy models to diverse platforms with \jing{different resource budgets} (\eg, latency, energy, memory constraints).
On the other hand, the computational budget varies due to the consumption of background apps that reduce the available computing capacity, and the energy budget varies due to the decreasing battery level of a mobile phone.
These scenarios give rise to a new challenge: \textit{how to instantly find an accurate and compact architectural configuration, given a specified superclass of interest and resource budget at runtime}.

To obtain compact architectures, most existing approaches either manually design \cite{howard2017mobilenets,sandler2018mobilenetv2} or use model compression \cite{molchanov2016pruning,zhuang2018discrimination,zhuang2018towards} and neural architecture search (NAS) \cite{NASRL2017,progressivedifferentiableNAS2019,cai2019proxylessnas} methods to find a specialized network. 
However, they are severely inflexible by a assumption that all the superclasses (each superclass corresponds to a group of classes) are predicted at the same time. In particular, for different \qdmy{superclass (\eg, animal and plant) classification tasks}, the optimal architectures can differ significantly, making this assumption undesirable to search for optimal architecture\qdmy{s} tailored for different task requirements. \jing{Repeatedly retraining the network to find architectures for each superclass} is extremely time-consuming and computation-intensive as the %
training cost grows linearly with the number of possible cases, and there are vast varieties of devices and dynamic deployment environments, which further exacerbate the efficiency problem significantly. 
%

To address the above issue, one may consider the one-shot NAS methods to decouple the model training and the architecture search stage, where
different sub-networks can be extracted from a well-trained over-parameterized network without any retraining~\cite{Cai2020Once_for_All,yu2020bignas}. However, existing methods suffer from two limitations. First, \jing{there exists a gap between training and deployment since the over-parameterized network is trained on all superclass data while specialized sub-networks focus on some specific superclasses. }
Second, existing one-shot NAS methods learn a surrogate model to estimate the model accuracy of the architecture and then use evolutionary algorithm~\cite{Cai2020Once_for_All} or coarse-to-fine strategy~\cite{yu2020bignas} 
to find optimal architectures during the deployment stage. Nevertheless, training a surrogate model requires plenty of architectures with the ground-truth accuracy
which are time-consuming to collect (\eg, 40 GPU hours to collect 16K sub-networks in OFA \cite{Cai2020Once_for_All}). Moreover, for $M$ requirements of different superclasses with diverse resources, existing methods have to search $M$ times,
which \jing{takes unbearable computational cost.}

To tackle these challenges, in this paper, we introduce a novel and general framework, called \methodfullname (\methodshortname), to deliver instant and effective trade-offs between accuracy and efficiency to support diverse superclasses of interest under different hardware constraints. Specifically, the proposed \methodshortname trains an over-parameterized network that supports many sub-networks and generates the architectures on the fly according to arbitrary deploying scenarios.
To this end, we propose a superclass dropout strategy that randomly drops the output logits corresponding to different superclasses during training. In this way, \jing{we enforce the over-parameterized network to pay more attention to the target superclass. As a result,} the over-parameterized network is robust to the superclass specification at testing time.
To enable specialization\qdmy{s} to arbitrary superclasses with different resources, we further propose an architecture generator to obtain architectures within a single forward pass, which is extremely efficient.

Our main contributions are summarized as follows:
\begin{itemize}[leftmargin=*]
\item We study a novel practical deep model deployment task for diverse superclasses of interest with arbitrary hardware resources, \jing{which has not been well studied.}
To resolve this, we devise a novel %
framework, 
called \methodfullname (\methodshortname), which supports versatile architectures for various deployment requirements at runtime.

\item 
We propose a superclass dropout strategy \jing{to pay more attention to the target superclass during training,}
making the over-parameterized network robust to the subsequent superclass specification during deployment.
We further design an architecture generator to generate architectures
without the need to search from scratch for each scenario, which greatly reduces the specialization cost.

\item
Experiments on three image classification datasets show the superior performance and efficiency of the proposed \methodshortname. For example, \jing{our EAS surpasses OFA by 1.3\% in Top-1 accuracy at the budget level of 361M \#MAdds on ImageNet-10. More impressively,} our proposed \methodshortname is able to find accurate and compact architectures within 0.1 second for 50 deployment scenarios.

\end{itemize}

\section{Related work}
\noindent\textbf{Network compression.} Network compression aims to reduce the model size and speed up the inference without significantly deteriorating the model performance. Existing network compression methods can be divided into several categories, including but not limited to network pruning~\cite{molchanov2016pruning,liu2018rethinking,guo2020dmcp,liudiscrimination2021}, network quantization~\cite{zhou2016dorefa,zhuang2018towards,esser2020learned,li2020additive}, knowledge distillation~\cite{romero2014fitnets,hinton2015distilling,zagoruyko2016paying,park2019relational}, low-rank decomposition~\cite{denton2014exploiting,jaderberg2014speeding,yu2017compressing,Peng_2018_ECCV}, \etc. 
Although achieving promising performance, prevailing methods only target predicting all the classes simultaneously. For different groups of classes, it requires repeatedly retraining
to find a compact network \qdmy{for each group}, which is time-consuming and inefficient. In contrast, our proposed \methodshortname trains an over-parameterized network that supports a \qdmy{great} number of sub-networks for diverse superclasses with different resources.

%
\noindent\textbf{Neural architecture search (NAS).}
NAS seeks to design efficient architectures automatically instead of relying on human expertise. Existing NAS methods can be roughly divided into three categories according to the search strategy, namely, reinforcement learning-based methods~\cite{NASRL2017,NASParametersharing2018,mnasnet2019,guo2020breaking}, evolutionary methods~\cite{real2017large,nsganet2019,real2019regularized,nsganetv22020}, and gradient-based methods~\cite{cai2019proxylessnas,DARTS2019,wu2019fbnet,sgnas}.
\jing{To efficiently explore the enormous search space, existing gradient-based multi-path NAS methods~\cite{wu2019fbnet,sgnas} relax the search space to be continuous and formulate \qdmy{the} optimization as a path selection problem. Compared with these methods, our method transforms the NAS problem into a subset selection problem following single-path NAS~\cite{stamoulis2019single,stamoulis2020single}, which significantly reduces the number of parameters and computational cost.}

When it comes to different resource constraints, existing methods have to repeatedly retrain the network and search for optimal architectures. To solve this, one-shot NAS methods~\cite{Cai2020Once_for_All,guo2020single,yu2020bignas} have been proposed to train a once-for-all network that supports different architectural configurations by decoupling the model training stage and the architecture search stage. To search for optimal architectures efficiently, one can use evolutionary search~\cite{Cai2020Once_for_All} or design an architecture generator to obtain effective architectures~\cite{randwire2019,generatorOptim2020}. However, given different resource constraints, these methods have to search architectures for each deployment scenario, which is inefficient.
Unlike these methods, we propose an efficient architecture generator that is able to obtain architectures on the fly within one forward pass given diverse superclasses with different resource constraints.

\noindent\textbf{Dynamic neural networks.} 
Dynamic neural networks, as opposed to static
ones, are able to adapt their structures or parameters to different inputs during inference and therefore enjoy the desired trade-off between accuracy and efficiency.
Existing methods can be roughly divided into three categories according to the granularity of dynamic networks, namely, instance-wise~\cite{cheng2020instanas,huang2017multi,lin2017runtime,wang2018skipnet}, spatial-wise~\cite{ren2018sbnet,wang2020glance,recasens2018learning} and temporal-wise dynamic networks~\cite{wu2019liteeval,fan2018watching,wu2020dynamic}. 
However, the inference efficiency of the data-dependent dynamic networks relies on the largest activated network in a batch of data, which limits the degree of practicality.
Moreover, 
all the above methods perform predictions \qdmy{on} all superclasses simultaneously.
Unlike these methods, our proposed \methodshortname is able to derive compact and accurate architectures given arbitrary superclasses and efficiency constraints.

\noindent\textbf{Multi-task learning.} Multi-task learning takes advantage of useful information contained in multiple related tasks to improve the performance of the learned models on all tasks. Existing methods can be roughly divided into several categories, including but not limited to feature learning approaches~\cite{zhang2014facial,liu2015multi}, low-rank methods~\cite{ando2005framework,chen2009convex}, task clustering approaches~\cite{thrun1996discovering,crammer2012learning}, task relation learning methods~\cite{evgeniou2004regularized,parameswaran2010large}, decomposition approaches~\cite{jalali2010dirty,chen2012learning}, \etc. Recently, several methods are proposed \qdmy{for architecture design~\cite{NEURIPS2020_634841a6} or fast specialization}~\cite{kim2019deep} under a limited number of budgets for multi-task learning.
Unlike these methods, our proposed method focuses on fast architecture specializations with elastic depth, width expansion ratio, and kernel size for diverse tasks under various resource budgets during deployment. More critically, most multi-task learning methods can be easily applied on top of our proposed method. For example, we can apply multi-task learning methods during the training of the over-parameterized network to improve the performance of all superclasses.

\begin{figure*}[t]
    \centering
    \begin{subfigure}[b]{0.34\textwidth}
         \centering
         \includegraphics[width=0.9\textwidth]{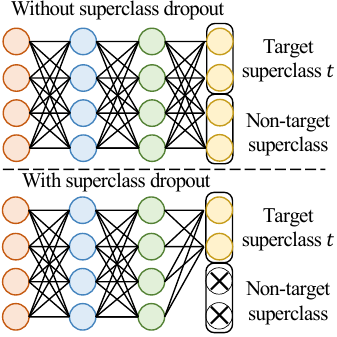}
         \caption{An illustration of superclass dropout}
         \label{fig:superclass_dropout}
    \end{subfigure}
    \hfill
    \begin{subfigure}[b]{0.6\textwidth}
         \centering
         \includegraphics[width=0.9\textwidth]{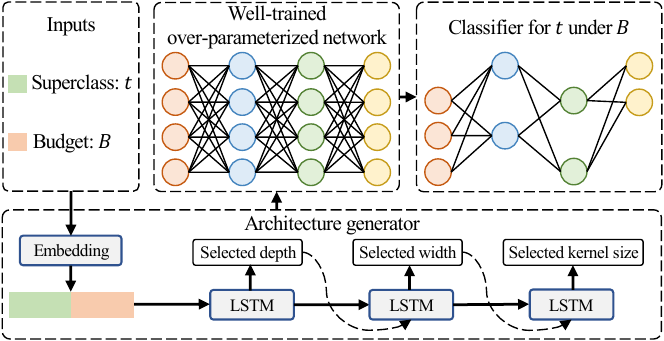}
         \caption{Architecture search for arbitrary superclasses and budgets}
         \label{fig:infer_EAG}
    \end{subfigure}
    \caption{
    The overview of the superclass dropout and architecture search. (a) We apply superclass dropout which randomly drops the output logits corresponding to different superclasses during the training of the over-parameterized network.
    (b) We build an architecture generator, which takes a superclass label $t$ and resource budget $B$ as inputs and generates architectures that satisfy the requirement. We use an embedding layer to map the superclass label and budget to the concatenated embedding vector
    }
    \label{fig:EAG}
\end{figure*}

\section{Elastic architecture search}
\noindent\textbf{Notation.} Let $\mD^t = \{ (\bx_n^t, y_n^t) \}_{n=1}^{N^t}$ be the training data for $t$-th superclass,
where $\bx_n^t \in \mX^t$, $y_n^t \in \mY^t$, and $N^t$ is the number of images. For simplicity, different superclasses of data are disjoint and focus on different classes, namely, $\forall i, j \in \{1, \dots, T\}$ and $i \neq j$, $\mY^i \cap \mY^j = \emptyset$, and $D^i \cap D^j = \emptyset$. 
Let $\mD = \{ (\bx_n, y_n) \}_{n=1}^N = \cup_{t=1}^T \mD^t $ be the union of data for $T$ superclasses, where %
$N=\sum_{t=1}^T{N^t}$.

In this paper, we focus on elastic architecture search problem that aims to obtain architectures given arbitrary superclasses with different resource budgets. To address this, one may use existing NAS methods~\cite{NASParametersharing2018,cai2019proxylessnas} to search architectures for each superclass and resource budget. However, when it comes to a vast amount of superclasses with various resource requirements, designing a specialized network for every scenario is computationally expensive and impractical.

To handle diverse superclasses with different resource budgets dynamically during testing, we propose a novel method, called \methodfullname (\methodshortname). 
Specifically, our proposed \methodshortname trains an over-parameterized network on $\mD$ that supports a very large number of sub-networks.
However, the training of the over-parameterized network is a highly entangled bi-level optimization problem~\cite{DARTS2019}. Inspired by one-shot NAS~\cite{Cai2020Once_for_All,yu2020bignas,guo2020single}, our solution is to decouple the model training stage and the architecture search stage. 
To make the model robust to the subsequent superclass specification during inference, we present a superclass dropout strategy at training time (See Section~\ref{sec:train_supernet}).
Once the over-parameterized network has been well-trained, we are able to
search architectures for different superclasses under various resources without any retraining.
In particular, we propose an efficient architecture generator to obtain optimal architectures within one forward pass (See Section~\ref{sec:train_generator}). 

\subsection{Superclass Dropout}
\label{sec:train_supernet}
Following~\cite{Cai2020Once_for_All}, the over-parameterized network consists of many sub-networks with different depths, width expansion ratios, and kernel sizes, where the weights of small sub-networks are shared with large sub-networks. To prevent interference between different sub-networks, we use the progressive shrinking strategy to train the over-parameterized network following~\cite{Cai2020Once_for_All}.
However, \jing{existing methods ignore the gap between the training and deployment where the over-parameterized network is trained on all superclass data while the specialized networks only concentrate on some specific superclasses.}
\jing{To mitigate the gap,}
one may train an over-parameterized network for each $\mD^t$ separately, which is computationally expensive and impractical. Such method also ignores that the data from other superclasses may contain useful context information 
for 
the target superclass, which may result in sub-optimal performance of the sub-networks.

To resolve these issues, motivated by~\cite{dropout2012,dropconnect2013}, we propose to train an over-parameterized network on $\mD$ with the superclass dropout strategy that randomly drops the output logits corresponding to non-target superclasses at each step of optimization, as shown in Figure~\ref{fig:superclass_dropout}.
Formally, let $\bv_n \in \mathbb{R}^{d}$ be the input feature of the last fully connected layer corresponding to $\bx_n$, where $d$ is the input feature dimension. For convenience, we omit the sample index $n$. The output logits of the network, denoted by $\bo$, can be computed by $\bo = \bw^\top \bv$,
where $\bw \in \mathbb{R}^{d \times C}$ denotes the weights of the last fully connected layer and $C$ is the number of classes. 
To encode the dropping decision of classes,
we introduce a binary mask $\bm\in \{0, 1\}^C$ to the output logits of the network:
\begin{equation}
    \label{eq:binary_mask}
    \hat{\bo} = \bm \odot \bo,
\end{equation}
where $\odot$ is an operation of the element-wise product and $\hat{\bo}$ is the masked version of $\bo$. Without any constraint, Eq.~(\ref{eq:binary_mask}) is a form of  class dropping. %
To enable superclass dropping,
the binary mask $\bm$ must be organized into groups by $\mG=\left\{G^{1}, \ldots, G^T \right\}$, where $\cup_{t=1}^{T} G^{t}=\{1, \ldots, C\}$, $G^i \cap G^j = \emptyset$ for $\forall i, j $ and $i \neq j$,  %
and $G^{t}$ 
denotes the index set of binary mask belonging to the $t$-th group. In this case, %
the output logits of the $t$-th superclass are 
\begin{equation}
    \label{eq:structural_binary_mask}
    \hat{\bo}_{G^t} = \bm_{G^t} \odot \bo_{G^t},
\end{equation}
where $\hat{\bo}_{G^t}$, $\bm_{G^t}$ and $\bo_{G^t}$ are the $t$-th group components of $\hat{\bo}$, $\bm$ and $\bo$, respectively. Here, the binary mask $\bm_{G^t}$ controls the dropping decision of the superclass $t$. 
During the model training stage, for any sample $(\bx^t, y^t) \in \mD^t$ in a batch of data, 
\jing{the binary mask $\bm_{G^t}$
corresponding to a non-target superclass $j$ (\ie, $\forall j \in \{ 1, \dots, T\}$ and $j \neq t$) is randomly set to zero} following a Bernoulli distribution with a hyper-parameter \ice{$q \in [0, 1]$} that controls the drop rate, \jing{while the binary mask corresponding to the target superclass $t$ is set to one.}
Then, we compute the output probabilit\qdmy{ies} using non-zero element\qdmy{s} in $\hat{\bo}_{G^t}$.
\jing{In this way, we enforce the over-parameterized network to pay more attention to the target superclass in each update step during training. As a consequence, the sub-networks derived from the over-parameterized network are more robust to superclass dropping during deployment.}

\subsection{Architecture Generator}
\label{sec:train_generator}
Once the over-parameterized network has been well-trained, we now focus on how to quickly find specialized sub-network\qdmy{s} from the over-parameterized network given arbitrary deployment scenarios.
To find optimal architectures, %
one may train a surrogate model to 
learn the mapping from an architectural configuration to its accuracy and then use \qdmy{the }evolutionary algorithm~\cite{Cai2020Once_for_All} or coarse-to-fine strategy~\cite{yu2020bignas} to explore the large configuration space. 
However, training such a surrogate model requires plenty of architecture-accuracy pairs that are very expensive to obtain in practice (\eg, 40 GPU hours to collect 16K sub-networks in OFA \cite{Cai2020Once_for_All}).
Moreover, for different superclasses with various resource budgets, 
existing methods have to search architectures for each deployment scenario,
which is extremely inefficient and unnecessary. 

To solve the above issue, we propose an architecture generator model to obtain architectures for diverse superclasses with different resources within a single forward pass.
As shown in Figure~\ref{fig:infer_EAG}, the architecture generator $G(B, t;\theta)$ is based on an LSTM network with three fully connected layers to predict the depth, width and kernel size configurations following~\cite{NASParametersharing2018}. 
Here, $\theta$ is the parameters of the architecture generator. 
Specifically, $G(B, t;\theta)$ takes a resource budget $B$ and superclass label $t$ as input\qdmy{s} and generates the architecture encoding $\alpha$.
Following~\cite{NASParametersharing2018,Guo2019NATNA}, we define $K$ budgets evenly sampled from the range of $[B_L, B_H]$, where $B_L$ and $B_H$ are the minimum and maximum of resource budgets, respectively.
To represent different superclasses and budgets, 
we build learnable embedding vectors for each of the superclass labels and predefined budgets. 
To deal with a resource budget with an arbitrary value, we use an embedding interpolation method following~\cite{guo2021pareto,radford2015unsupervised}. 
We then concatenate the resulting budget and superclass embedding vectors and feed them into the LSTM network. 
By incorporating the learnable embedding vectors into the parameters of the architecture generator, we are able to train them jointly. 
The training details of the proposed architecture generator are shown in Algorithm~\ref{alg:generator_training}. 

\begin{algorithm}[t]
	\caption{Training method for the architecture generator.}
    	\begin{algorithmic}[1]\small
    		\REQUIRE The parameters of the well-trained over-parameterized network $\bW$, minimum of the resource budget $B_L$, maximum of the resource budget $B_H$, uniform distribution $U(B_L, B_H)$, total superclass number $T$, superclass index set $\{1, \dots, T\}$, learning rate $\eta$, training data set $D=\cup_{t=1}^T \mD^t $, hyper-parameters $\lambda$ and $\tau$.
    		\STATE Initialize the architecture generator parameters $\theta$.
    		\WHILE{not converged}
    		    \STATE Randomly select a superclass label $t$ from $\{1, \dots, T\}$.
    		    \STATE Sample a batch of data from $\mD^t$.
    		    \STATE Sample a resource budget $B$ from $U(B_L, B_H)$.
    		    \STATE Compute the loss using Eq. (\ref{eq:joint_loss}).
    		    \STATE Update the architecture generator by~\\
                \STATE ~~~~~~~~~$\theta \leftarrow \theta - \eta \nabla_{\theta} \mL(\theta)$. \\
    		\ENDWHILE \\
    	\end{algorithmic}
		\label{alg:generator_training}
\end{algorithm}

\noindent{\textbf{Objective function.}}
The goal of the architecture generator is to obtain an architecture encoding $\alpha$ corresponding to a sub-network \qdmy{of} the well-trained over-parameterized network that minimizes the validation loss $\mL_{\mathrm{val}}(\bW, \alpha)$,
where $\bW$ is the weights of the %
over-parameterized network. Here, the validation loss is the cross-entropy on the validation set. To ensure that the computational cost $R(\alpha)$ is close to $B$, we introduce a computational constraint loss, which can be formulated as
\begin{equation}
    \mL_C(\alpha, B) = (R(\alpha) - B)^2.
\end{equation}
By considering both the validation loss and computational constraint loss, the joint loss function can be formulated as
\begin{equation}
    \label{eq:joint_loss}
    \mL(\theta) =  \mL_{\mathrm{val}}(\bW, G(B, t;\theta)) + \lambda \mL_C(G(B, t;\theta), B) ,
\end{equation}
where $\lambda$ is a hyper-parameter that balances two loss terms. 

\noindent{\textbf{Single-path architecture encoding.}}
Finding a sub-network from 
the well-trained over-parameterized network is a discrete and non-differentiable process.
To address this, we propose to use single-path framework
to encode architectural configurations (\ie, kernel size, width expansion ratio) in the search space of the well-trained over-parameterized network and formulate the NAS problem as a subset selection problem.
In this case, the architecture encoding $\alpha$ can be represented by a series of binary gates that determine which subset of weights to use and the computational cost $R(\alpha)$ is a function of the binary gates.

\jing{To encode kernel size and width expansion ratio configurations, we use the architecture encoding following single-path NAS~\cite{stamoulis2020single}. Motivated by this, we also propose to encode the depth configurations in a single-path scheme.} 
Specifically,
a shallower network can be viewed as \qdmy{a} subset of layers of the whole network. Thus, the depth configuration can also be determined by a set of binary gates. Formally, let $\mH^l(\bx)$ be the mapping until the $l$-th module and $\mF(\cdot)$ be the mapping of the $(l+1)$-th module. Note that a module can be a layer or a block that consists of a collection of successive layers. Then, the mapping $\mH^{l+1}(\bx)$ until the $(l+1)$-th module can be formulated as
\begin{equation}
    \label{eq:depth_decision}
    \mH^{l+1}(\bx) = \mH^{l}(\bx) \cdot (1 - g^{l+1}) + \mF(\mH^{l}(\bx)) \cdot g^{l+1},
\end{equation}
where $g^{l+1}$ is a binary gate to determine whether to go through the $(l+1)$-th module
during the forward propagation. 
In this case, the computational cost $R^{l+1}$ until the $(l+1)$-th module is
\begin{equation}
    R^{l+1} = R^{l} + r^{l+1} \cdot g^{l+1},
\end{equation}
where $r^{l+1}$ is the computational cost of the $(l+1)$-th module.

Instead of determining the output of the binary gates using some heuristic metrics (\eg, the $\ell_1$-norm of the weights), we force the output of each binary gate following the Bernoulli distribution with a probability $p_i$, where $i$ denotes the index of the binary gate.
Formally, the output of the $i$-th binary gate can be computed by
\begin{equation}
    \label{eq:bernoulli}
    g(p_i) =\left\{\begin{array}{ll}
    1 & \text { with probability } p_i, \\
    0 & \text { with probability } 1 - p_i.
    \end{array}\right.
\end{equation}
Therefore, the architecture encoding $\alpha$ is encoded by a set of binary gates and each binary gate is sampled from the Bernoulli distribution.
To obtain $p_i$, we apply a sigmoid function $\sigma(\cdot)$ to the output $\bz$ of the LSTM network, which can be formulated as
\begin{equation}
    p_i = \sigma(z_i),
\end{equation}
where $z_i$ is the $i$-th element of $\bz$.

\noindent{\textbf{Gumbel-Softmax for differentiable relaxation.}}
To train the architecture generator, we need to estimate the gradient of the objective function $\mL(\theta)$ \wrt~the probability $p_i$. However, Eq.~(\ref{eq:bernoulli}) is not differentiable \wrt ~$p_i$.
Following~\cite{li2020differentiable}, we use the Gumbel-Softmax reparameterization trick~\cite{maddison2016concrete,jang2016categorical} to reparameterize $p_i$ as
\begin{equation}
    \label{eq:backward}
    h(p_i, \tau) = \sigma\left(\left(\log \frac{p_i}{1-p_i}+\log \frac{u_i}{1-u_i}\right) / \tau\right),
\end{equation}
where $u_i \sim U(0,1)$ is the random noise sampled from the uniform distribution $U(0,1)$ and $\tau$ is the temperature hyper-parameter.
During the forward propagation, the output of a binary gate is computed by
\begin{equation}
    \label{eq:forward}
    \hat{g}(p_i) = 
    \begin{cases} 
    1 & \text {if}~h(p_i,\tau) > 0.5, \\
    0 & \text {otherwise},
    \end{cases}
\end{equation}
where $\hat{g}(p_i)$ is the hard version of $h(p_i, \tau)$. 
During backward propagation, we use straight-through estimator (STE)~\cite{bengio2013estimating,zhou2016dorefa} to approximate the gradient of $\hat{g}(p_i)$ by the gradient of $h(p_i,\tau)$.
In this way, the objective function $\mL(\theta)$ is differentiable \wrt~$p_i$. Therefore, we are able to train the architecture generator using gradient-based methods.

\newcommand{\ofa}{OFA-V\xspace}
\newcommand{\ofat}{OFA-T\xspace}

\begin{table*}[t]
	\caption{Performance comparisons on ImageNet-10. We report the Top-1 Accuracy (Acc.) of different architectures on diverse superclasses. ``Avg. Acc.'' and ``Avg. \#MAdds'' denote the average Top-1 accuracy and the average number of multiply-adds, respectively. \jing{A column of “T-t” shows the Top-1 accuracy on the t-th superclass. Different row sets report the Top-1 accuracy at different budget levels}}
	\label{tab:imagenet_t_10}
 \begin{center}
 \begin{threeparttable}
    \resizebox{0.95\textwidth}{!}{
 	\begin{tabular}{c||cccccccccc|c|c}
 	    \hline %
        Method & T-1 & T-2 & T-3 & T-4 & T-5 & T-6 & T-7 & T-8 & T-9 & T-10 & \tabincell{c}{Avg. Acc. (\%)} & \tabincell{c}{Avg. \#MAdds (M)} \\
        \hline\hline %
        \ofa  & 86.7 & 90.7 & 79.7 & 81.0 & 85.0 & \textbf{89.0} & 82.0 & 96.0 & 92.7 & \textbf{96.7} & 87.9 & 230  \\
        \ofat & 86.3 & 90.0 & 79.7 & 79.0 & 85.3 & 87.3 & \textbf{83.0} & \textbf{97.0} & 93.3 & \textbf{96.7} & 87.8 & 230 \\
        \methodshortname (Ours) & \textbf{89.0} & \textbf{91.3} & \textbf{81.7} & \textbf{81.3} & \textbf{86.7} & 88.7 & \textbf{83.0} & 96.7 & \textbf{94.0} & \textbf{96.7} & \textbf{88.9} & \textbf{219} \\
        \hline %
        \ofa  & 87.7 & 91.0 & 80.0 & 80.0 & 84.7 & 89.7 & 83.0 & 95.7 & 93.7 & 95.3 & 88.0 & 278 \\
        \ofat & 88.0 & 90.0 & 80.0 & 80.0 & 86.7 & 89.3 & \textbf{85.0} & 95.7 & \textbf{94.3} & 95.7& 88.5& 279 \\
        \methodshortname (Ours) & \textbf{90.0} & \textbf{92.0} & \textbf{81.3} & \textbf{80.3} & \textbf{87.7} & \textbf{90.3} & 84.3 & \textbf{97.3} & \textbf{94.3} & \textbf{96.7}& \textbf{89.4} & \textbf{269} \\
        \hline %
        \ofa  & 88.3 & 90.7 & 81.0 & 79.7 & 85.7 & \bf{90.3} & 83.7 & 95.0 & 92.7 & \textbf{96.7} & 88.4 & 329  \\
        \ofat & 88.0 & 90.0 & 80.0 & 80.3 & 86.0 & 90.0 & \textbf{85.7} & 95.7 & 93.7 & 95.7& 88.5& 328 \\
        \methodshortname (Ours) & \textbf{89.3} & \textbf{91.7} & \textbf{82.0} & \textbf{82.3} & \textbf{89.0} & 90.0 & 84.3 & \textbf{97.3} & \textbf{94.3} & \textbf{96.7} & \textbf{89.7} & \textbf{316} \\
        \hline %
        \ofa  & 87.0 & 90.0 & 80.7 & 79.7& 85.7 & 90.0 & 83.7 & 95.0 & 93.0 & 96.7 & 88.1 & 364 \\
        \ofat & 87.3 & 91.3 & 80.0 & 80.7 & 87.0 & 89.3 & 84.7 & 96.3 & 93.3 & 96.3& 88.6 & 373 \\
        \methodshortname (Ours) & \textbf{88.7} & \textbf{91.7} & \textbf{82.0} & \textbf{82.3} & \textbf{88.7} & \textbf{90.3} & \textbf{85.3} & \textbf{97.3} & \textbf{95.0} & \textbf{97.3} & \textbf{89.9} & \textbf{361} \\
        \hline %
	\end{tabular}}
	 \end{threeparttable}
	 \end{center}
	 \vspace{-0.2in}
\end{table*}

\begin{table*}[t]
    \setlength{\abovecaptionskip}{0.cm}
	\caption{Performance comparisons on ImageNet-12. We report the Top-1 Accuracy (Acc.) of different architectures on diverse superclasses. ``Avg. Acc.'' and ``Avg. \#MAdds'' denote the average Top-1 accuracy and the average number of multiply-adds, respectively. \jing{A column of “T-t” shows the Top-1 accuracy on the t-th superclass. Different row sets report the Top-1 accuracy at different budget levels}}
    \setlength{\belowcaptionskip}{-0.cm}
	\label{tab:imagenet_t_12}
 \begin{center}
 \begin{threeparttable}
    \resizebox{1.\textwidth}{!}{
 	\begin{tabular}{c||cccccccccccc|c|c}
 	    \hline %
        Method & T-1 & T-2 & T-3 & T-4 & T-5 & T-6 & T-7 & T-8 & T-9 & T-10 & T-11 & T-12 & \tabincell{c}{Avg. Acc. (\%)} & \tabincell{c}{Avg. \#MAdds (M)} \\
        \hline\hline %
        \ofa  & 81.7 & 86.6 & 92.0 & 85.8 & 88.1 & 77.3 & 83.4 & 79.5 & 80.3 & 84.7 & 79.5 & \textbf{79.4} & 83.2 & 229 \\
        \ofat & 80.9 & 86.6 & 92.5 & 85.6 & 88.1 & 77.4 & 83.4 & \textbf{79.6} & 80.8 & \textbf{85.2} & 79.7 & 79.1 & 83.2 & 229 \\
        \methodshortname (Ours) & \textbf{81.8} & \textbf{86.9} & \textbf{93.1} & \textbf{85.9} & \textbf{88.5} & \textbf{78.4} & \textbf{83.5} & \textbf{79.6} & \textbf{80.9} & 84.7 & \textbf{79.8} & 78.1 & \textbf{83.4} & \textbf{223} \\
        \hline %
        \ofa  & \textbf{81.7} & 87.0 & 92.8 & \textbf{86.6} & 87.8 & 78.0 & 83.7 & 78.9 & 81.6 & 84.9 & \textbf{80.5} & 78.6 & 83.5 & 278 \\
        \ofat & 81.1 & 87.0 & 92.6 & 86.2 & \textbf{89.1} & \textbf{79.1} & \textbf{83.8} & \textbf{79.5} & 81.3 & 85.2 & 80.2 & 78.5 & 83.6 & 279 \\
        \methodshortname (Ours) & 81.3 & \textbf{87.4} & \textbf{92.9} & \textbf{86.6} & 88.1 & 77.8 & 83.3 & 79.3 & \textbf{81.9} & \textbf{85.5} & 80.3 & \textbf{80.0} & \textbf{83.7} & \textbf{266} \\
        \hline %
        \ofa & 81.2 & \textbf{87.9} & 92.7 & 85.9 & 88.4 & 78.3 & 84.1 & \textbf{79.5} & 81.7 & 84.9 & \textbf{80.8} & 78.8 & 83.7 & 329\\
        \ofat & 81.1 & 87.2 & \textbf{93.0} & 86.2 & 89.0 & \textbf{79.3} & 83.9 & 79.1 & 81.1 & 85.2 & 80.5 & 78.8 & 83.7 & 329 \\
        \methodshortname (Ours) & \textbf{81.6} & 87.5 & \textbf{93.0} & \textbf{86.5} & \textbf{89.1} & 78.6 & \textbf{84.3} & 79.3 & \textbf{82.3} & \textbf{85.3} & 80.1 & \textbf{79.6} & \textbf{83.9} & \textbf{320} \\
        \hline %
        \ofa & 81.0 & 87.2 & 92.8 & 85.9 & 88.5 & 78.3 & 84.4 & \textbf{80.2} & 81.5 & 85.0 & \textbf{81.1} & 78.7 & 83.7 & 379\\
        \ofat & 80.8 & 87.3 & \textbf{93.3} & 85.9 & 89.0 & \textbf{79.6} & \textbf{84.6} & 79.9 & 81.7 & 85.4 & 80.0 & 79.0 & 83.9 & 379 \\
        \methodshortname (Ours) & \textbf{81.5} & \textbf{87.7} & \textbf{93.3} & \textbf{86.8} & \textbf{89.4} & 79.2 & 84.2 & \textbf{80.2} & \textbf{82.0} & \textbf{86.0} & 80.9 & \textbf{79.9} & \textbf{84.3} & \textbf{358} \\
        \hline %
	\end{tabular}}
	 \end{threeparttable}
	 \end{center}
	 	 \vspace{-0.2in}
\end{table*}
\noindent{\textbf{Inferring architectures for diverse superclasses with different resources.}}
Once the architecture generator has been \jing{well-}trained, we can instantly search sub-networks that meet specified deployment scenarios.
Specifically, given a superclass label and resource budget, the output of the architecture generator is the probabilities of the binary gates encoding the architectural configurations. 
\jing{Since we consider computational constraint loss during training, the architectures sampled from the learned distribution are expected to satisfy the computational constraint. 
}
We will repeat the sampling process if the sampled ones violate the resource budget. Last, we select the final architecture with the highest validation accuracy. Moreover, 
our architecture generator is able to generate architectures for $M$ deployment requirements within a single forward pass by setting the mini-batch size to $M$.

\section{Experiments}
\label{sec:experiments}
\noindent{\textbf{Datasets.}}
We evaluate the proposed \methodshortname on the large-scale image classification dataset, namely ImageNet~\cite{deng2009imagenet}. We also evaluate our method on \qdmy{the} fine-grained classification dataset Stanford Cars~\cite{Krause20133DOR}. \jing{For each dataset, we construct superclasses by grouping semantically similar classes.}
Based on ImageNet, we construct two datasets following~\cite{robustness}. The first dataset, denoted by ImageNet-10, consists of 10 superclasses and each superclass contains 6 classes. The total number of training and testing samples for ImageNet-10 are 77,237 and 3,000, respectively. The second dataset, denoted by ImageNet-12, consists of 12 superclasses and each superclass contains 20 classes. The total number of training and testing samples for ImageNet-12 are 308,494 and 12,000, respectively. \qdmy{Based on Stanford Cars, we construct the third dataset, denoted by Car-20, which consists of 20 superclasses and each superclass contains 3 classes. The total number of training and testing samples are 2,494 and 2,463, respectively.} For ImageNet-10 and ImageNet-12 datasets, we randomly choose 10k training samples as the validation set. For Car-20 dataset, we randomly choose 500 training samples as the validation set. More details about the datasets are put in the supplementary material.

\noindent{\textbf{Search space.}} We apply our proposed \methodshortname to MobileNetV3~\cite{howard2019searching} search space. Specifically, the model is split into 5 units following~\cite{Cai2020Once_for_All}.
We choose the depth of each unit from $\{ 2, 3, 4 \}$, the width expansion ratio of each inverted residual block from $\{ 3, 4, 6 \}$, and the kernel size of each depthwise convolution from $\{ 3, 5, 7 \}$. 

\noindent{\textbf{Compared methods.}} We compare our \methodshortname with 
the state-of-the-art one-shot NAS method OFA~\cite{Cai2020Once_for_All}.
To perform classification on different superclasses, we construct the following variants for comparisons. \textbf{\ofa}: the vanilla OFA that trains an over-parameterized network and only performs a single evolutionary search for all superclasses. \textbf{\ofat}: trains an over-parameterized network and searches optimal architectures for each superclass. 

\noindent{\textbf{Evaluation metrics.}} We measure the performance of different methods using the Top-1 accuracy on diverse superclasses given a specific budget. Following~\cite{sandler2018mobilenetv2,howard2019searching}, we measure the computational cost of the searched architectures using the number of multiply-adds (\#MAdds). Following~\cite{DARTS2019}, we also use the training cost on a GPU device (GeForce RTX 3090) to measure the time of training an architecture generator and the search cost on a CPU device (Intel Xeon Gold 6230R) to measure the time of using the generator to find an optimal architecture. 

\noindent{\textbf{Implementation details.}}
We apply our proposed \methodshortname to MobileNetV3~\cite{howard2019searching} search space following~\cite{Cai2020Once_for_All}.
We first train an over-parameterized network on all superclasses of training data for 120 epochs using the progressive shrinking strategy~\cite{Cai2020Once_for_All} with a mini-batch size of 256. Different from OFA~\cite{Cai2020Once_for_All}, we share the kernel weights among different kernel sizes without using the kernel transformation. 
We use SGD with a momentum of 0.9 for optimization~\cite{qian1999momentum}. The learning rate starts at 0.01 and decays with cosine
annealing~\cite{loshchilov2016sgdr}. We set the drop rate
$q$ to 0.6, 0.1, and 0.15 for the experiments on ImageNet-10, ImageNet-12, and Car-20, respectively. We then train the architecture generator for $90$ epochs on the validation set. The number of learnable embedding vectors $K$ for the resource budget is set to $10$. We set the dimension of the embedding vector to 32 for both the superclass label and resource budget. We use Adam for optimization~\cite{Kingma2015AdamAM} with a learning rate of $1 \times 10^{-3}$. The hyper-parameters $\lambda$ and $\tau$ are set to $0.01$ and $1$, respectively. For all experiments on ImageNet-10, ImageNet-12 \qdmy{and Car-20}, we set $B_L$ and $B_H$ to $150$ and $550$, respectively. We put more implementation details in the supplementary material.

\begin{figure}[ht]
    \centering
    \begin{minipage}[t]{0.48\textwidth}
        \centering
        \includegraphics[width=0.9\linewidth]{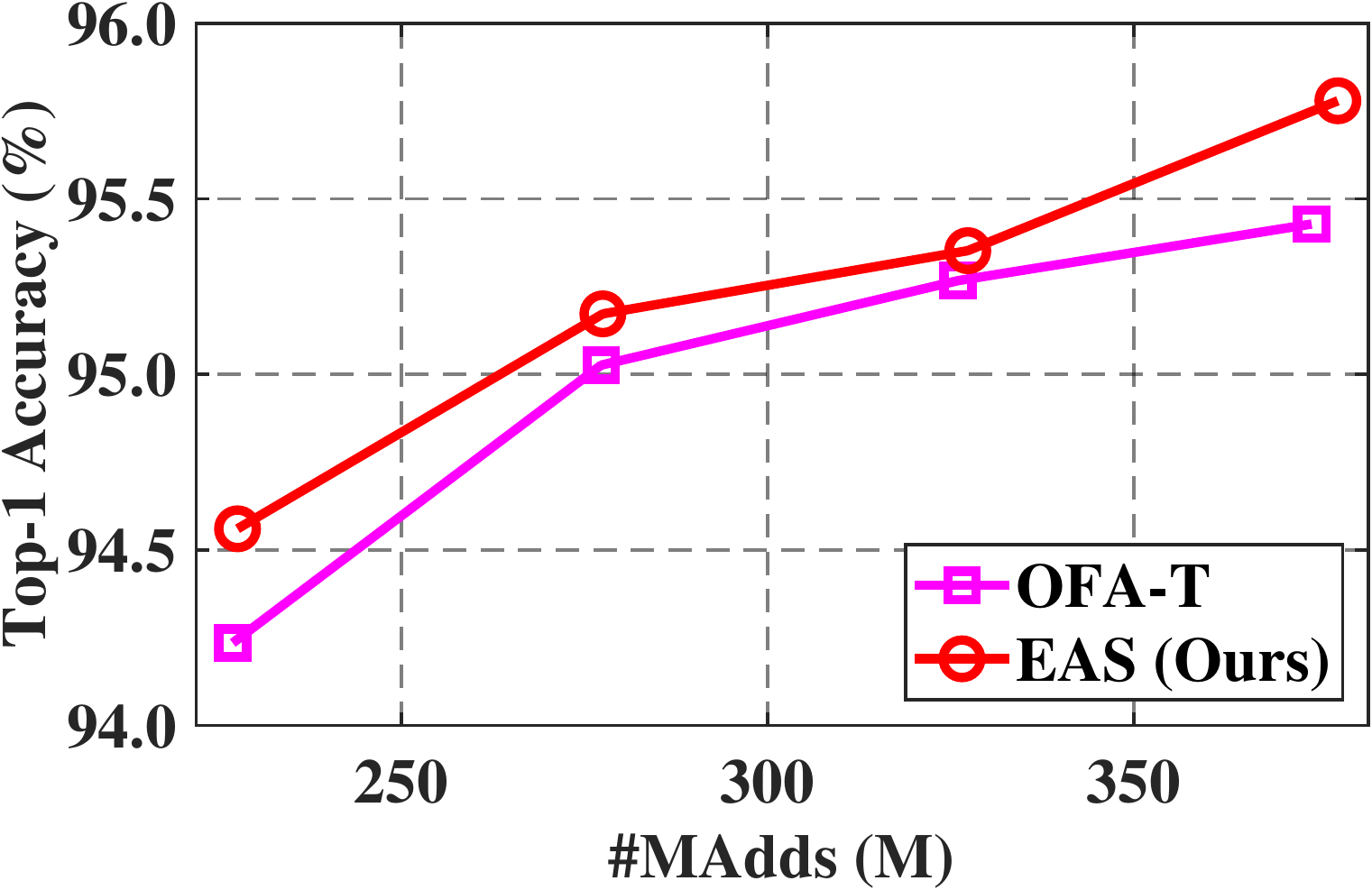}
        \caption{Performance comparisons of the proposed \methodshortname with \ofat under various budget levels on Car-20}
        \label{fig:effect_superclass_dropout_on_car20}
    \end{minipage}
    \hfill
    \begin{minipage}[t]{0.48\textwidth}
        \centering
        \includegraphics[width=0.9\linewidth]{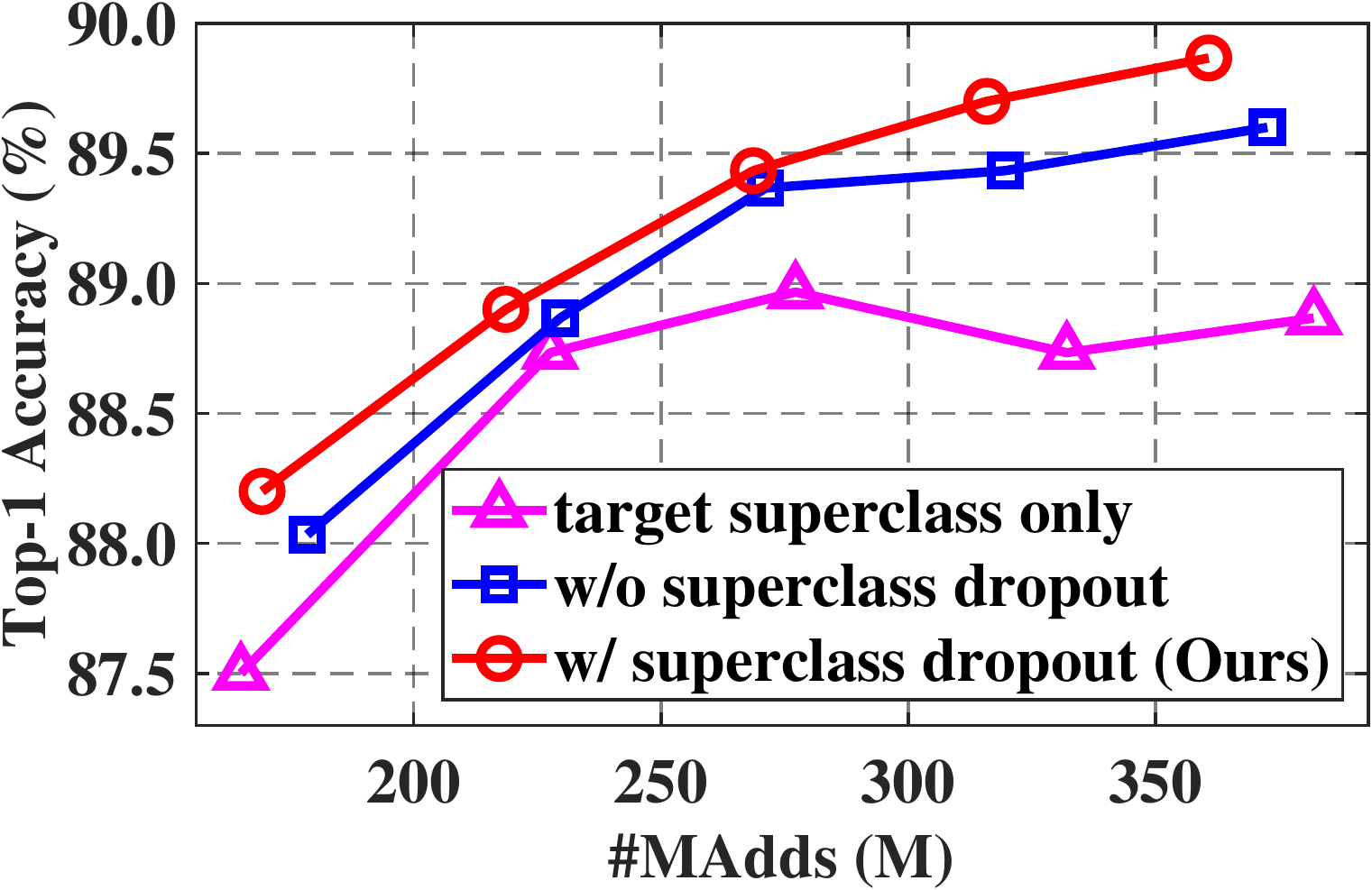}
        \caption{Performance comparisons of the proposed \methodshortname with different superclass dropout strategies on ImageNet-10}
        \label{fig:effect_superclass_dropout}
    \end{minipage}
    \vspace{-0.2in}
\end{figure}
\begin{table*}[t]
    \centering
    \caption{
    Comparisons of the training time and search time for architecture search among different methods on ImageNet-10. The training time for \ofat is the cost of sampling architecture-accuracy pairs and learning a surrogate model. For the multi-path AG and single-path AG, the training time is the cost of learning an architecture generator. The search time is the cost of finding architectures for 10 superclasses and 5 resource budgets
    }
    \begin{threeparttable}
    \resizebox{0.8\textwidth}{!}
    {      
    \begin{tabular}{c||cccc} 
    \hline %
    Method &  \ofat & Multi-path AG & Single-path AG (Ours) \\ 
    \hline\hline %
    Training Time (GPU hour) & 104.0 & \textbf{40.3} & \textbf{0.9}\\
    Search Time (CPU second) & 3474.1 & \textbf{$<$0.1} & \textbf{$<$0.1} \\
    \hline %
    \end{tabular}
    }
    \end{threeparttable}
    \label{tab:generation_cost}
\end{table*}


\subsection{Main Results}
We show the results \qdmy{on ImageNet-10 and ImageNet-12} in Table~\ref{tab:imagenet_t_10} and Table~\ref{tab:imagenet_t_12}\qdmy{, respectively. We also report the average Top-1 accuracy and average \#MAdds across all superclasses of Car-20 in Figure~\ref{fig:effect_superclass_dropout_on_car20}.} We put more results on Car-20 and visualization of the generated architectures in the supplementary material. 
Compared with \ofa, \ofat achieves better or comparable performance in all superclasses with different resource budgets. For example, on ImageNet-10,
\ofat surpasses \ofa by 0.5\% in terms of the average model accuracy at the resource budget level of 373M Avg. \#MAdds. These results demonstrate that searching architectures for all superclasses may not be optimal for each superclass. Compared with \ofat, our method achieves better or comparable model accuracy in all superclasses with diverse resource budgets. For the average performance of all superclasses on different resource budgets, our \methodshortname consistently outperforms \ofat \jing{by a large margin} with fewer or comparable \#MAdds.
For example, on ImageNet-10, our \methodshortname surpasses \ofat by 1.3 \% on the average Top-1 accuracy at the budget level of 361M Avg. \#MAdds.
These results verify the effectiveness of our method.
Note that for superclass-2 of ImageNet-10, the obtained architecture of \methodshortname achieves the best accuracy at the resource budget level of 269M Avg. \#MAdds. With the increase of \#MAdds, the accuracy of the searched architecture goes worse. Similar results are also observed for the baseline methods, such as superclass-7 of ImageNet-10 for \ofat and superclass-1 of ImageNet-12 for \ofa. 
One possible reason is that for those superclasses of data that are easy to be classified, using a compact network with low computational overhead is sufficient to obtain promising performance. Increasing the resource budget may suffer from the over-fitting issue.


\subsection{Ablation Studies}
\noindent{\textbf{Effectiveness of the superclass dropout.}}
To investigate the effect of the superclass dropout strategy, we first train the over-parameterized networks with and without superclass dropout on ImageNet-10 and then use the proposed architecture generator to obtain architectures. We also train the over-parameterized network with the target superclass only by setting the drop rate $q$ to 1, which indicates that we completely drop the output logits corresponding to the non-target superclasses during training. More results in terms of different drop rates are put in the supplementary material.
We report the average Top-1 accuracy and average \#MAdds across all superclasses in Figure~\ref{fig:effect_superclass_dropout}. From the results, \methodshortname without superclass dropout surpasses the one with the target superclass only, especially at high resource budget settings. These results verify that exploiting useful context information in other superclasses improves the performance of current superclass. More critically, \methodshortname with superclass dropout consistently outperforms the one without superclass dropout and the performance improvement brought by the superclass dropout strategy is more significant at higher resource budget levels. For example, the architecture searched by the superclass dropout strategy yields a much lower average \#MAdds (316M vs. 373M) but achieves better performance.
These results demonstrate the effectiveness of the proposed superclass dropout strategy.

\begin{figure}[t]
    \centering
    \includegraphics[width=0.9\linewidth]{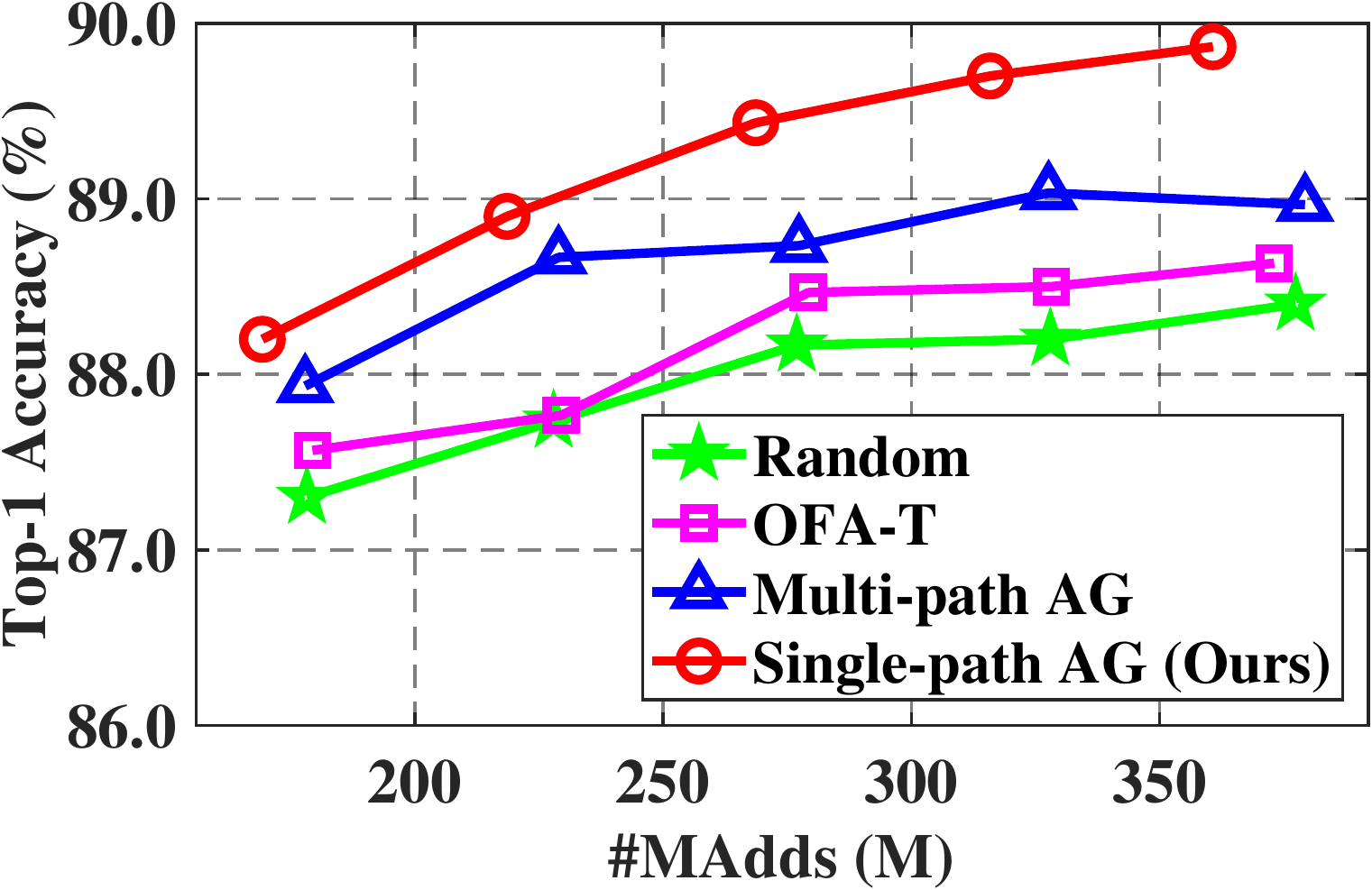}
    \caption{Performance comparisons of the proposed \methodshortname with different search algorithms on ImageNet-10
    }
    \label{fig:effect_architecture_generator}
    \label{-0.3in}
\end{figure}

\noindent{\textbf{Effectiveness of the architecture generator.}}
To investigate the effect of the architecture generator, we first train the over-parameterized network with superclass dropout on ImageNet-10 and then use the following methods to search \qdmy{for} the architectures. 1) \textbf{Random}: randomly sample\qdmy{s an} architecture for each deployment scenario.
2) \textbf{\ofat}: uses the evolutionary algorithm to search optimal architectures for diverse superclasses and resources~\cite{Cai2020Once_for_All}. 3) \textbf{Multi-path AG}: based on the proposed architecture generator, we use the multi-path scheme~\cite{DARTS2019,wu2019fbnet} to encode the generated architectures for each superclass. 4) \textbf{Single-path AG}: uses the proposed architecture generator to obtain architectures with the single-path architecture encoding for each superclass. 
We report the average Top-1 accuracy and average \#MAdds across all superclasses in Figure~\ref{fig:effect_architecture_generator}. We also show the training cost and search cost in Table~\ref{tab:generation_cost}. 
From the results, all the methods consistently outperform Random under various levels of \#MAdds. 
The multi-path AG consistently outperforms \ofat at different levels of \#MAdds. Our proposed single-path AG further outperforms the multi-path scheme by a large margin (89.43\% vs. 88.73\% at the budget level of 277M Avg. \#MAdds), which demonstrates the effectiveness of the proposed architecture generator with the single-path architecture encoding. 
Compared with \ofat, the multi-path AG has orders of magnitude lower training cost and search cost. 
More critically, our single-path AG further reduces the training cost, which demonstrates the efficiency of our single-path architecture generator in both training and architecture search. For example, our proposed architecture generator is able to generate architectures within 0.1 second for 50 development requirements.

\section{Conclusion}
\label{sec:conclusion}
In this paper, we have studied a new challenging problem of efficient deployment for different superclasses with various resources, where the superclass of interest and resource constraint are dynamically specified at testing time. To this end, we have proposed a novel and general framework, called \methodfullname (\methodshortname), to instantly specialize neural networks at runtime. 
The proposed \methodshortname first trains an over-parameterized network with the \textbf{superclass dropout} strategy that randomly drops the output logits corresponding to different superclasses. In this way, the over-parameterized network supports many sub-networks for diverse superclasses once trained. Based on the well-trained over-parameterized network, we have further proposed an efficient architecture generator to instantly obtain architectures within a single forward pass given arbitrary superclasses and budgets during inference. Experiments on image classification datasets have shown the superiority of the proposed methods 
under diverse deployment scenarios.


\section{Limitations and Future Work}
The proposed EAS may bring some negative societal impacts, such as job losses for algorithm engineers. Moreover, improper uses of the proposed \methodshortname may result in disastrous effects. For example, obtaining a very small network for an autonomous driving car may bring greater risks to the driver.

In the future, we may extend our method in three aspects. First, we currently perform neural architecture search based on the given specific superclass and budget. However, the superclass label is often unknown in many real-world applications. To address this, we will need to perform superclass prediction first and then use the proposed method to obtain architectures. Second, we may consider multiple superclasses at testing time, \jing{where different superclass combinations might have overlapped classes. This is a very}
challenging scenario due to the large combinatorial space. For example, for 100 superclasses, we have roughly $2^{100}$ different superclass combinations. Third, we may extend our proposed method to broader applications, such as dense prediction, machine translation, \etc.

{\small
\bibliographystyle{ieee_fullname}
\bibliography{egbib}
}

\newcommand{\easv}{EAS-V\xspace}
\onecolumn
{
\centering
\large

\textbf{Appendix} \\
\vspace{1.0em}
}
\appendix

\renewcommand\thesection{\Alph{section}}
\renewcommand\thefigure{\Alph{figure}}
\renewcommand\thetable{\Alph{table}}
\renewcommand{\theequation}{\Alph{equation}}

\setcounter{table}{0}
\setcounter{figure}{0}

We organize our supplementary material as follows. 
\begin{itemize}[leftmargin=*]
    \item In Section~\ref{sec:details_dataset}, we provide more details about the datasets.
    \item In Section~\ref{sec:implementation_details}, we describe more implementation details about the proposed method and baselines.
    \qdmy{\item In Section~\ref{sec:results_car_20}, we show the results of different methods on Car-20 dataset.}
    \item In Section~\ref{sec:effect_drop_rates}, we investigate the effect of different drop rates.
    \qdmy{\item In Section~\ref{sec:easv_vs_eas}, we include more results to investigate the effect of searching architectures for each superclass.
    } 
    \item In Section~\ref{sec:visualization}, we show the visualization results of the searched architectures.
\end{itemize}

\section{More details about the datasets}
\label{sec:details_dataset}
In this section, we introduce more details about the datasets mentioned in Section 4. Based on ImageNet~\cite{deng2009imagenet}, we construct two datasets for different superclasses following~\cite{robustness}, namely ImageNet-10 and ImageNet-12. \qdmy{Based on Stanford Cars, we construct Car-20.} We show more details of ImageNet-10, ImageNet-12 and Car-20 in Tables~\ref{tab:imagenet10_details}, ~\ref{tab:imagenet12_details} and ~\ref{tab:stanford_cars20_details}, respectively. 

\begin{table*}[!ht]
\centering
\caption{More details about ImageNet-10. We report the names of superclass\qdmy{es} and classes in it. Each class name is separated by a comma.}
\label{tab:imagenet10_details}
\renewcommand\arraystretch{1.1}
\renewcommand{\tabcolsep}{5.0pt}
\scalebox{0.9}{
\begin{tabular}{c|c}
\hline
Superclass & Class \\
\hline \hline
Dog & Chihuahua, Japanese spaniel, Maltese dog, Pekinese, Shih-Tzu, Blenheim spaniel\\
Bird & Cock, Hen, Ostrich, Brambling, Goldfinch, House finch \\
Insect & Tiger beetle, Ladybug, Ground beetle, Long-horned beetle, Leaf beetle, Dung beetle \\
Monkey & Guenon, Patas, Baboon, Macaque, Langur, Colobus \\
Car & Jeep, Limousine, Cab, Beach wagon, Ambulance, Convertible \\
Cat & Leopard, Snow leopard, Jaguar, Lion, Cougar, Lynx \\
Truck & Tow truck, Moving van, Fire engine, Pickup, Garbage truck, Police van \\
Fruit & Granny Smith, Rapeseed, Corn, Acorn, Hip, Buckeye \\
Fungus & Agaric, Gyromitra, Stinkhorn, Earthstar, Hen-of-the-woods, Coral fungus \\
Boat & Gondola, Fireboat, Speedboat, Lifeboat, Yawl, Canoe \\
\hline      
\end{tabular}
}
\end{table*}

\begin{table*}[!h]
\centering
\caption{More details about ImageNet-12. We report the names of superclass\qdmy{es} and classes in it. Each class name is separated by a comma.}
\label{tab:imagenet12_details}
\renewcommand\arraystretch{1.1}
\renewcommand{\tabcolsep}{5.0pt}
\scalebox{0.93}{
\begin{tabular}{c|c}
\hline
Superclass & Class \\
\hline \hline
Dog & \makecell[c]{Chihuahua, Japanese spaniel, Maltese dog, Pekinese, Shih-Tzu, \\Blenheim spaniel, Papillon, Toy terrier, Rhodesian ridgeback, Afghan hound, \\Basset, Beagle, Bloodhound, Bluetick, Black-and-tan coonhound, \\Walker hound, English foxhound, Redbone, Borzoi, Irish wolfhound}\\
\hline
Structure & \makecell[c]{Dam, Altar, Dock, Apiary, Bannister, \\Barbershop, Barn, Beacon, Boathouse, Bookshop, \\Brass, Breakwater, Butcher shop, Castle, Chainlink fence, \\Church, Cinema, Cliff dwelling, Coil, Confectionery}\\
\hline
Bird & \makecell[c]{Cock, Hen, Ostrich, Brambling, Goldfinch, \\House finch, Junco, Indigo bunting, Robin, Bulbul, \\Jay, Magpie, Chickadee, Water ouzel, Kite, \\Bald eagle, Vulture, Great grey owl, African grey, Macaw}\\
\hline
Clothing & \makecell[c]{Cowboy hat, Crash helmet, Abaya, Academic gown, Diaper, \\Apron, Feather boa, Football helmet, Bathing cap, Bearskin, \\Fur coat, Bikini, Gown, Bolo tie, Bonnet, \\Bow tie, Brassiere, Hoopskirt, Cardigan, Christmas stocking}\\
\hline
Vehicle & \makecell[c]{Minivan, Model T, Ambulance, Amphibian, Electric locomotive, \\Fire engine, Barrow, Forklift, Beach wagon, Freight car, \\Garbage truck, Bicycle-built-for-two, Go-kart, Half track, Cab, \\Horse cart, Jeep, Jinrikisha, Limousine, Convertible}\\
\hline
Reptile & \makecell[c]{Loggerhead, Leatherback turtle, Mud turtle, Terrapin, Box turtle, \\Banded gecko, Common iguana, American chameleon, Whiptail, Agama, \\Frilled lizard, Alligator lizard, Gila monster, Green lizard,  African chameleon, \\Komodo dragon, African crocodile, American alligator, Triceratops, Thunder snake}\\
\hline
Carnivore & \makecell[c]{Timber wolf, White wolf, Red wolf, Coyote, Dingo, \\Dhole, African hunting dog, Hyena, Red fox, Kit fox, \\Arctic fox, Grey fox, Cougar, Lynx, Leopard, \\Snow leopard, Jaguar, Lion, Tiger, Cheetah}\\
\hline
Insect & \makecell[c]{Tiger beetle, Ladybug, Ground beetle, Long-horned beetle, Leaf beetle, \\Dung beetle, Rhinoceros beetle, Weevil, Fly, Bee, \\Ant, Grasshopper, Cricket, Walking stick, Cockroach, \\Mantis, Cicada, Leafhopper, Lacewing, Dragonfly}\\
\hline
Instrument & \makecell[c]{Cornet, Maraca, Marimba, Accordion, Acoustic guitar, \\Drum, Electric guitar, Banjo, Oboe, Ocarina, \\Flute, Organ, Bassoon, French horn, Gong, \\Grand piano, Harmonica, Harp, Cello, Chime}\\
\hline
Food & \makecell[c]{French loaf, Bagel, Pretzel, Head cabbage, Broccoli, \\Cauliflower, Zucchini, Spaghetti squash, Acorn squash, Butternut squash, \\Cucumber, Artichoke, Bell pepper, Cardoon, Mushroom, \\Strawberry, Orange, Lemon, Fig, Pineapple}\\
\hline
Furniture & \makecell[c]{Cradle, Crib, Medicine chest, Desk, Dining table, \\Entertainment center, Barber chair, File, Bassinet, Folding chair, \\Four-poster, Studio couch, Park bench, Bookcase, Table lamp, \\Throne, Toilet seat, Chiffonier, China cabinet, Rocking chair}\\
\hline
Primate & \makecell[c]{Indri, Orangutan, Gorilla, Chimpanzee, Gibbon, \\Siamang, Guenon, Patas, Baboon, Macaque, \\Langur, Colobus, Proboscis monkey, Marmoset, Capuchin, \\Howler monkey, Titi, Spider monkey, Squirrel monkey, Madagascar cat} \\
\hline      
\end{tabular}
}
\end{table*}

\section{More implementation details}
\label{sec:implementation_details}
In this section, we introduce more implementation details mentioned in Section 4.
Following~\cite{Cai2020Once_for_All,yu2019universally}, we use the knowledge distillation technique~\cite{hinton2015distilling} to train the over-parameterized network. Specifically, we take the largest sub-network from the over-parameterized network as the teacher network. We train the teacher network for 120 epochs with a mini-batch size of 256. We use SGD with a momentum of 0.9 for optimization~\cite{qian1999momentum}. The weight decay is set to $3 \times 10^{-5}$.
The learning rate starts at 0.1 and decays with cosine annealing~\cite{loshchilov2016sgdr}. We do not use the superclass dropout strategy during teacher training. For \ofa and \ofat, we first sample 16K sub-networks with different architectures and measure their accuracies of diverse superclasses on the validation set following~\cite{Cai2020Once_for_All}. We then train the accuracy predictor for 250 epochs using a mini-batch size of 256. We use SGD with momentum for optimization. The momentum term and weight decay are set to 0.9 and
$1 \times 10^{-4}$, respectively. The learning rate is initialized
to 0.1 and decreased to 0 following the cosine function. 

\begin{table*}[!t]
    \centering
    \caption{More details about Car-20. We report the names of superclass\qdmy{es} and classes in it. Each class name is separated by a comma.}
    \label{tab:stanford_cars20_details}
    \renewcommand\arraystretch{1.1}
    \renewcommand{\tabcolsep}{5.0pt}
    \scalebox{0.83}{
    \begin{tabular}{c|c}
    \hline
    Superclass & Class \\
    \hline \hline
    Acura & Integra Type R 2001, TL Type-S 2008, ZDX Hatchback 2012 \\
    Aston Martin & V8 Vantage Coupe 2012, Virage Convertible 2012, Virage Coupe 2012 \\
    Audi & R8 Coupe 2012, S4 Sedan 2012, S5 Coupe 2012 \\
    Bentley & Arnage Sedan 2009, Continental Flying Spur Sedan 2007, Continental Supersports Conv. Convertible 2012 \\
    BMW & 3 Series Sedan 2012, M6 Convertible 2010, Z4 Convertible 2012 \\
    Buick & Enclave SUV 2012, Rainier SUV 2007, Verano Sedan 2012 \\
    Chevrolet & Cobalt SS 2010, Silverado 1500 Classic Extended Cab 2007, TrailBlazer SS 2009 \\
    Chrysler & PT Cruiser Convertible 2008, Sebring Convertible 2010, Town and Country Minivan 2012 \\
    Dodge & Challenger SRT8 2011, Magnum Wagon 2008, Ram Pickup 3500 Quad Cab 2009 \\
    Ferrari & 458 Italia Convertible 2012, Italia Coupe 2012, California Convertible 2012 \\
    Ford & F-450 Super Duty Crew Cab 2012, GT Coupe 2006, Mustang Convertible 2007 \\
    GMC & Acadia SUV 2012, Savana Van 2012, Terrain SUV 2012 \\
    Honda & Accord Sedan 2012, Odyssey Minivan 2007, Odyssey Minivan 2012 \\
    Hyundai & Accent Sedan 2012, Tucson SUV 2012, Veloster Hatchback 2012 \\
    Jeep & Grand Cherokee SUV 2012, Liberty SUV 2012, Wrangler SUV 2012 \\
    Lamborghini & Aventador Coupe 2012, Gallardo LP 570-4 Superleggera 2012, Reventon Coupe 2008 \\
    Mercedes-Benz & 300-Class Convertible 1993, SL-Class Coupe 2009, Sprinter Van 2012 \\
    Nissan & Juke Hatchback 2012, Leaf Hatchback 2012, NV Passenger Van 2012 \\
    Suzuki & Aerio Sedan 2007, SX4 Hatchback 2012, SX4 Sedan 2012 \\
    Toyota & Camry Sedan 2012, Corolla Sedan 2012, Sequoia SUV 2012 \\
    \hline      
    \end{tabular}
    }
\end{table*}

\begin{table*}[!t]
    \centering
    \setlength{\abovecaptionskip}{0.cm}
    \caption{
    Performance comparisons on Car-20. We report the Top-1 Accuracy (Acc.) of different architectures on diverse superclasses. ``Avg. Acc.'' and ``Avg. \#MAdds'' denote the average Top-1 accuracy and the average number of multiply-adds, respectively.
    }
    \setlength{\belowcaptionskip}{-0.cm}
    \renewcommand\arraystretch{1.1}
    \renewcommand{\tabcolsep}{5.0pt}
    \label{tab:Car_t_20}
    \begin{center}
    \begin{threeparttable}
    \resizebox{1.\textwidth}{!}{
    \begin{tabular}{c||cccccccccccccccccccc|c|c} 
    \hline %
    Method & T-1 & T-2 & T-3 & T-4 & T-5 & T-6 & T-7 & T-8 & T-9 & T-10 & T-11 & T-12 & T-13 & T-14 & T-15 & T-16 & T-17 & T-18 & T-19 & T-20 & \tabincell{c}{Avg. Acc. (\%)} & \tabincell{c}{Avg. \#MAdds (M)} \\ 
    \hline\hline %
    \ofat & 97.6 & \textbf{87.5} & \textbf{84.7} & \textbf{95.1} & \textbf{95.8} & 98.3 & \textbf{100.0} & \textbf{100.0} & \textbf{100.0} & 80.8 & \textbf{99.2} & 99.3 & \textbf{96.7} & \textbf{88.0} & 97.0 & 88.6 & \textbf{100.0} & \textbf{100.0} & \textbf{81.7} & 94.4 & 94.2 & \textbf{227.0} \\ 
    \methodshortname (Ours) & \textbf{98.4} & 86.6 & \textbf{84.7} & \textbf{95.1} & 95.0 & \textbf{100.0} & 97.6 & 99.2 & \textbf{100.0} & \textbf{83.3} & \textbf{99.2} & \textbf{100.0} & 95.9 & \textbf{88.0} & \textbf{98.5} & \textbf{93.0} & 99.2 & \textbf{100.0} & \textbf{81.7} & \textbf{96.0} & \textbf{94.6} & 227.6 \\ \hline
    \ofat & 97.6 & \textbf{87.5} & 85.5 & \textbf{95.1} & \textbf{97.5} & \textbf{100.0} & \textbf{100.0} & \textbf{100.0} & \textbf{100.0} & 83.3 & 98.5 & 99.3 & 96.7 & \textbf{90.7} & \textbf{100.0} & 88.6 & 99.2 & \textbf{100.0} & \textbf{85.0} & \textbf{96.0} & 95.0 & \textbf{277.3} \\ 
    \methodshortname (Ours) & \textbf{99.2} & \textbf{87.5} & \textbf{87.9} & 92.7 & 95.0 & 99.2 & \textbf{100.0} & \textbf{100.0} & \textbf{100.0} & \textbf{85.0} & \textbf{99.2} & \textbf{100.0} & \textbf{97.5} & 88.9 & \textbf{100.0} & \textbf{91.2} & \textbf{100.0} & \textbf{100.0} & \textbf{85.0} & 95.2 & \textbf{95.2} & 277.5 \\ \hline
    \ofat & 97.6 & \textbf{88.4} & \textbf{87.9} & \textbf{95.1} & \textbf{98.3} & 99.2 & \textbf{100.0} & \textbf{100.0} & \textbf{100.0} & 83.3 & \textbf{99.2} & \textbf{100.0} & 95.9 & 91.7 & \textbf{100.0} & 88.6 & \textbf{100.0} & \textbf{100.0} & \textbf{84.2} & \textbf{96.0} & 95.3 & \textbf{326.0} \\ 
    \methodshortname (Ours) & \textbf{98.4} & 84.8 & \textbf{87.9} & \textbf{95.1} & 95.8 & \textbf{100.0} & \textbf{100.0} & \textbf{100.0} & \textbf{100.0} & \textbf{86.7} & \textbf{99.2} & \textbf{100.0} & \textbf{97.5} & \textbf{92.6} & 99.2 & \textbf{91.2} & \textbf{100.0} & \textbf{100.0} & \textbf{84.2} & 94.4 & \textbf{95.4} & 327.3 \\ \hline
    \ofat & 98.4 & \textbf{88.4} & \textbf{88.7} & 95.1 & \textbf{97.5} & \textbf{100.0} & \textbf{100.0} & \textbf{100.0} & \textbf{100.0} & \textbf{85.0} & 98.5 & \textbf{100.0} & \textbf{95.9} & 91.7 & \textbf{100.0} & 87.7 & \textbf{100.0} & \textbf{100.0} & 84.2 & \textbf{97.6} & 95.4 & \textbf{374.3} \\ 
    \methodshortname (Ours) & \textbf{99.2} & 87.5 & 87.9 & \textbf{96.7} & 96.6 & 99.2 & \textbf{100.0} & \textbf{100.0} & \textbf{100.0} & \textbf{85.0} & \textbf{99.2} & \textbf{100.0} & \textbf{95.9} & \textbf{94.4} & \textbf{100.0} & \textbf{92.1} & \textbf{100.0} & \textbf{100.0} & \textbf{85.0} & 96.8 & \textbf{95.8} & 377.8 \\ \hline
    \end{tabular}}
    \end{threeparttable}
    \end{center}
\end{table*}


\qdmy{
\section{More results on Car-20}
\label{sec:results_car_20}
In this section, we provide more results on the fine-grained dataset Car-20. We show the results of different methods in Table~\ref{tab:Car_t_20}. For the average performance of all superclasses on different resource budgets, our \methodshortname consistently outperforms \ofat in terms of the model accuracy with comparable \#MAdds. For example, our method surpasses \ofat by 0.4\% at the resource budget level of 377.8 M Avg. \#MAdds. These results verify the effectiveness of our method.
}

\section{Effect of different drop rates}
\label{sec:effect_drop_rates}
\begin{figure}[!tb]
    \centering
    \begin{minipage}{.46\textwidth}
        \centering
        \includegraphics[width=0.9\linewidth]{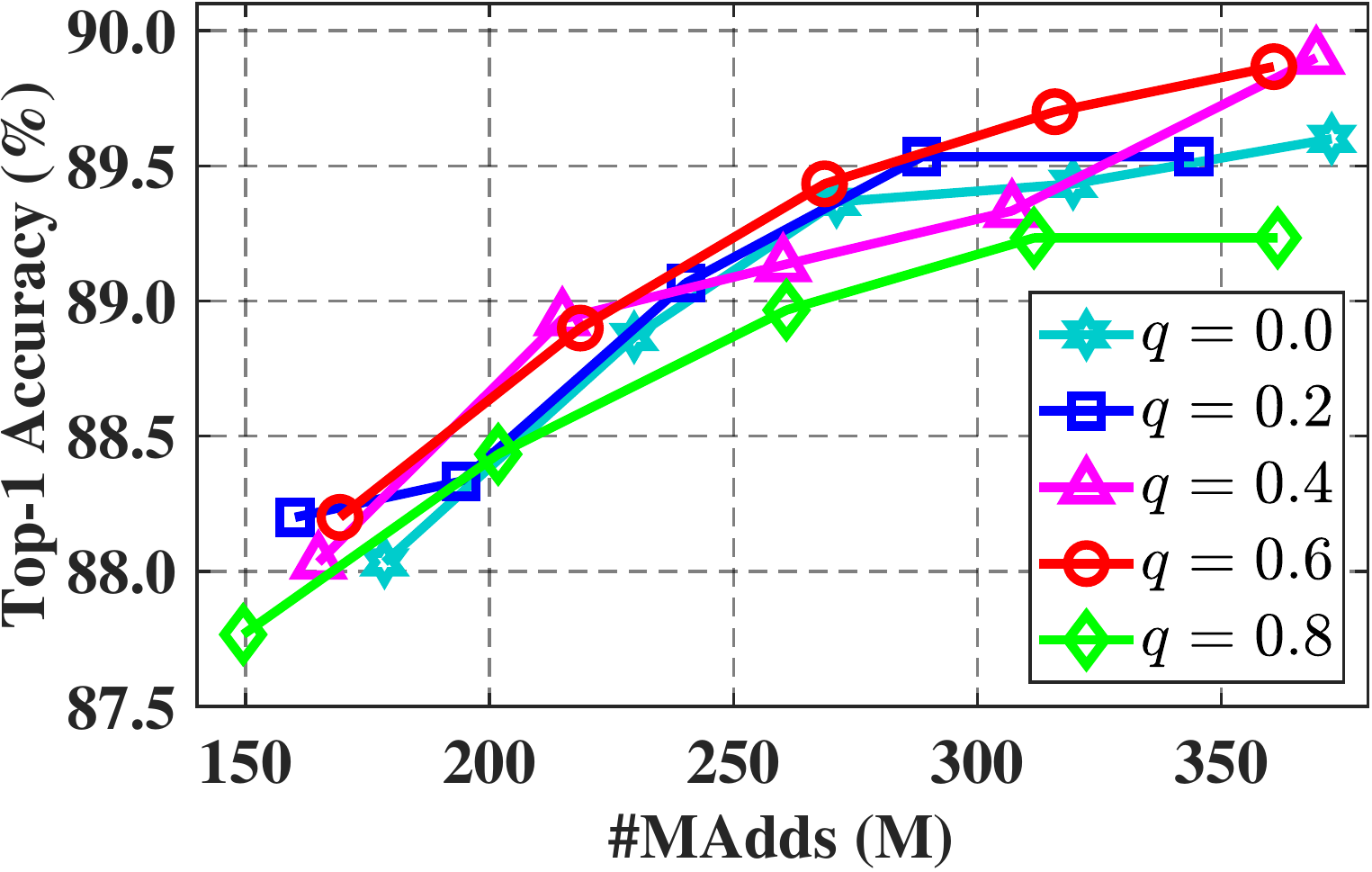}
        \caption{Comparisons of the proposed \methodshortname with different drop rates on ImageNet-10.}
    \label{fig:effect_drop_rate}
    \end{minipage}%
    \hfill
    \begin{minipage}{0.46\textwidth}
        \centering
        \includegraphics[width=0.9\linewidth]{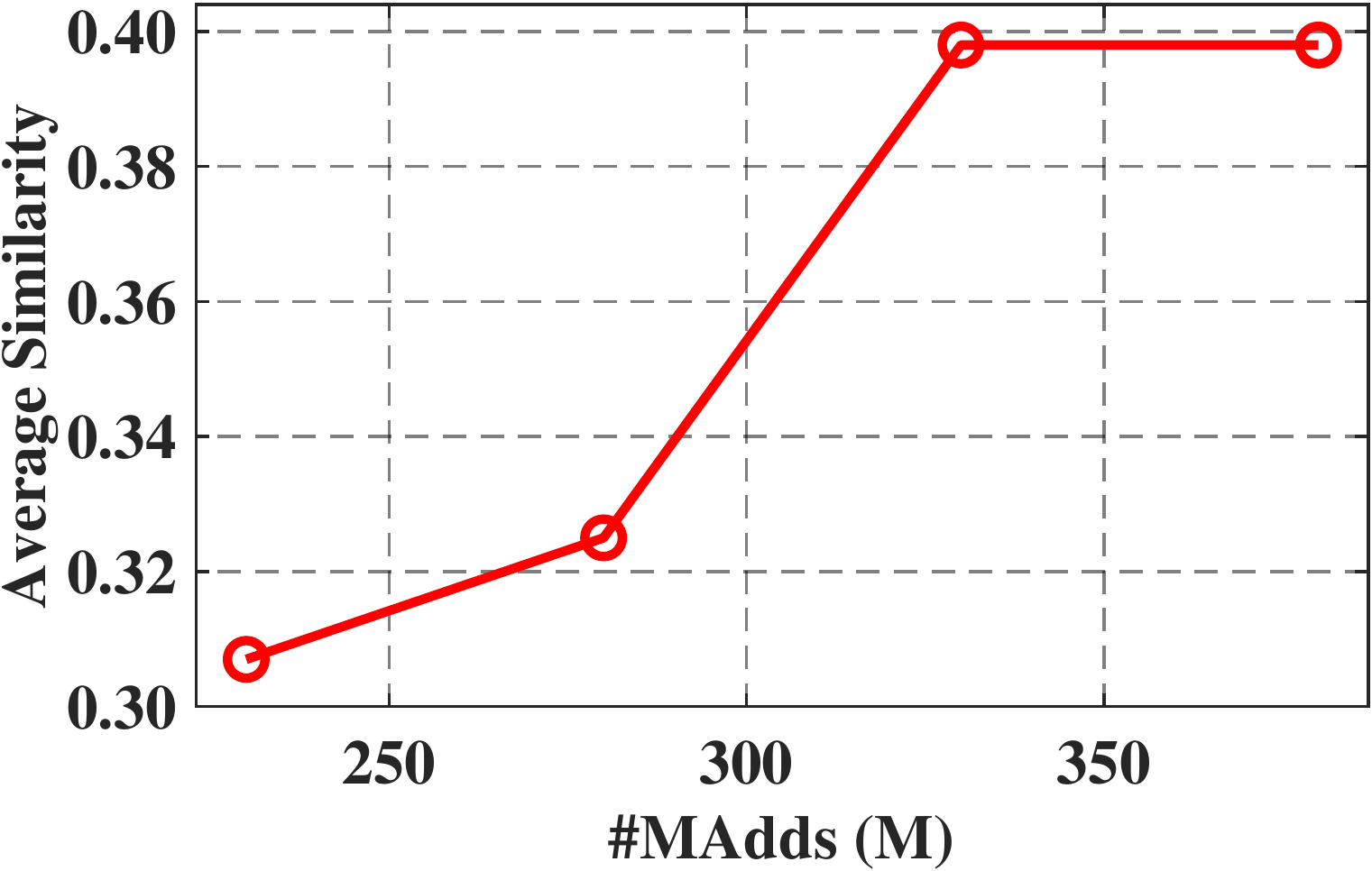}
        \caption{The average architecture cosine similarity \vs~\#MAdds on ImageNet-10.
        }
        \label{fig:simi_arch}
    \end{minipage}
\end{figure}

To investigate the effect of the superclass dropout strategy, we first train the over-parameterized networks with different drop rates $q$ on ImageNet-10 and then use the proposed architecture generator to obtain architectures. We report the results in Figure~\ref{fig:effect_drop_rate}. From the results, 
\methodshortname with a drop rate of 0.6 achieves the best performance. For example, at the \#MAdds level of 316 M, the Top-1 accuracy of the proposed \methodshortname with a drop rate of 0.6 is 89.7\%, which is higher than those with other drop rates. Hence, we set $q$ to 0.6 by default in our experiments on ImageNet-10. Moreover, with the increase of $q$, the performance of the searched architectures first goes better and then goes worse. These results demonstrate the effectiveness of the proposed superclass dropout strategy. 

\section{More results on the effectiveness of searching architectures for each superclass}
\label{sec:easv_vs_eas}
To further investigate the effectiveness of searching architectures \qdmy{for} each superclass, we first train the over-parameterized network with superclass dropout on ImageNet-10 and then use the following methods to search \qdmy{for} architectures. \textbf{\methodshortname-V}: we apply the proposed \methodshortname to find a single architecture for all superclasses. \textbf{\methodshortname}: we use the proposed \methodshortname to find an architecture for each superclass.
From Table~\ref{tab:easv_vs_eas_img10}, finding a specific architecture for each superclass outperforms that of searching a single architecture for all superclasses. For example, at the average \#MAdds level of 219 M, \methodshortname outperforms \methodshortname-V by 0.8\% in terms of the average Top-1 accuracy. These results show the necessity of finding architectures for each superclass.

\begin{table*}[t]
    \setlength{\abovecaptionskip}{0.cm}
	\caption{Performance comparisons between \easv and \methodshortname on ImageNet-10. \easv denotes that we apply the proposed \methodshortname to find a single architecture for all superclasses. \methodshortname indicates that we apply the proposed \methodshortname to find an architecture for each superclass. We report the Top-1 Accuracy (Acc.) of different architectures on diverse superclasses. ``T-$t$'' indicates the $t$-th superclass.  ``Avg. Acc.'' and ``Avg. \#MAdds'' denote the average Top-1 accuracy and the average number of multiply-adds, respectively.}
    \setlength{\belowcaptionskip}{-0.cm}
	\label{tab:easv_vs_eas_img10}
 \begin{center}
 \begin{threeparttable}
    \resizebox{0.75\textwidth}{!}{
 	\begin{tabular}{c||cccccccccc|c|c}
 	    \hline %
        Method & T-1 & T-2 & T-3 & T-4 & T-5 & T-6 & T-7 & T-8 & T-9 & T-10 & \tabincell{c}{Avg. Acc. (\%)} & \tabincell{c}{Avg. \#MAdds (M)} \\
        \hline\hline %
        \easv  & 87.3 & \textbf{91.3} & 79.3 & \textbf{81.3} & 83.7 & \textbf{88.7} & \textbf{86.7} & 95.0 & 93.0 & 95.0 & 88.1 & 223  \\
        \methodshortname (Ours) & \textbf{89.0} & \textbf{91.3} & \textbf{81.7} & \textbf{81.3} & \textbf{86.7} & \textbf{88.7} & 83.0 & \textbf{96.7} & \textbf{94.0} & \textbf{96.7} & \textbf{88.9} & \textbf{219} \\
        \hline %
        \easv  & 88.0 & 91.3 & \textbf{81.7} & 79.0 & 84.7 & \textbf{90.7} & 82.3 & 96.3 & 93.7 & \textbf{96.7} & 88.4 & 274 \\
        \methodshortname (Ours) & \textbf{90.0} & \textbf{92.0} & 81.3 & \textbf{80.3} & \textbf{87.7} & 90.3 & \textbf{84.3} & \textbf{97.3} & \textbf{94.3} & \textbf{96.7}& \textbf{89.4} & \textbf{269} \\
        \hline %
        \easv  & 88.7 & 91.0 & 80.0 & 80.3 & 86.0 & 88.7 & \textbf{84.7} & 96.0 & 93.7 & 96.3 & 88.5 & 320  \\
        \methodshortname (Ours) & \textbf{89.3} & \textbf{91.7} & \textbf{82.0} & \textbf{82.3} & \textbf{89.0} & \textbf{90.0} & 84.3 & \textbf{97.3} & \textbf{94.3} & \textbf{96.7} & \textbf{89.7} & \textbf{316} \\
        \hline %
        \easv  & \textbf{89.0} & 91.3 & 81.7 & 80.0 & 87.0 & 88.0 & 82.7 & 95.3 & 94.3 & 96.7 & 88.6 & 374 \\
        \methodshortname (Ours) & 88.7 & \textbf{91.7} & \textbf{82.0} & \textbf{82.3} & \textbf{88.7} & \textbf{90.3} & \textbf{85.3} & \textbf{97.3} & \textbf{95.0} & \textbf{97.3} & \textbf{89.9} & \textbf{361} \\
        \hline %
	\end{tabular}}
	 \end{threeparttable}
	 \end{center}
\end{table*}

\section{Visualization of the searched architectures}
\label{sec:visualization}
\begin{figure}[!b]
    \centering
    \includegraphics[width=0.95\linewidth]{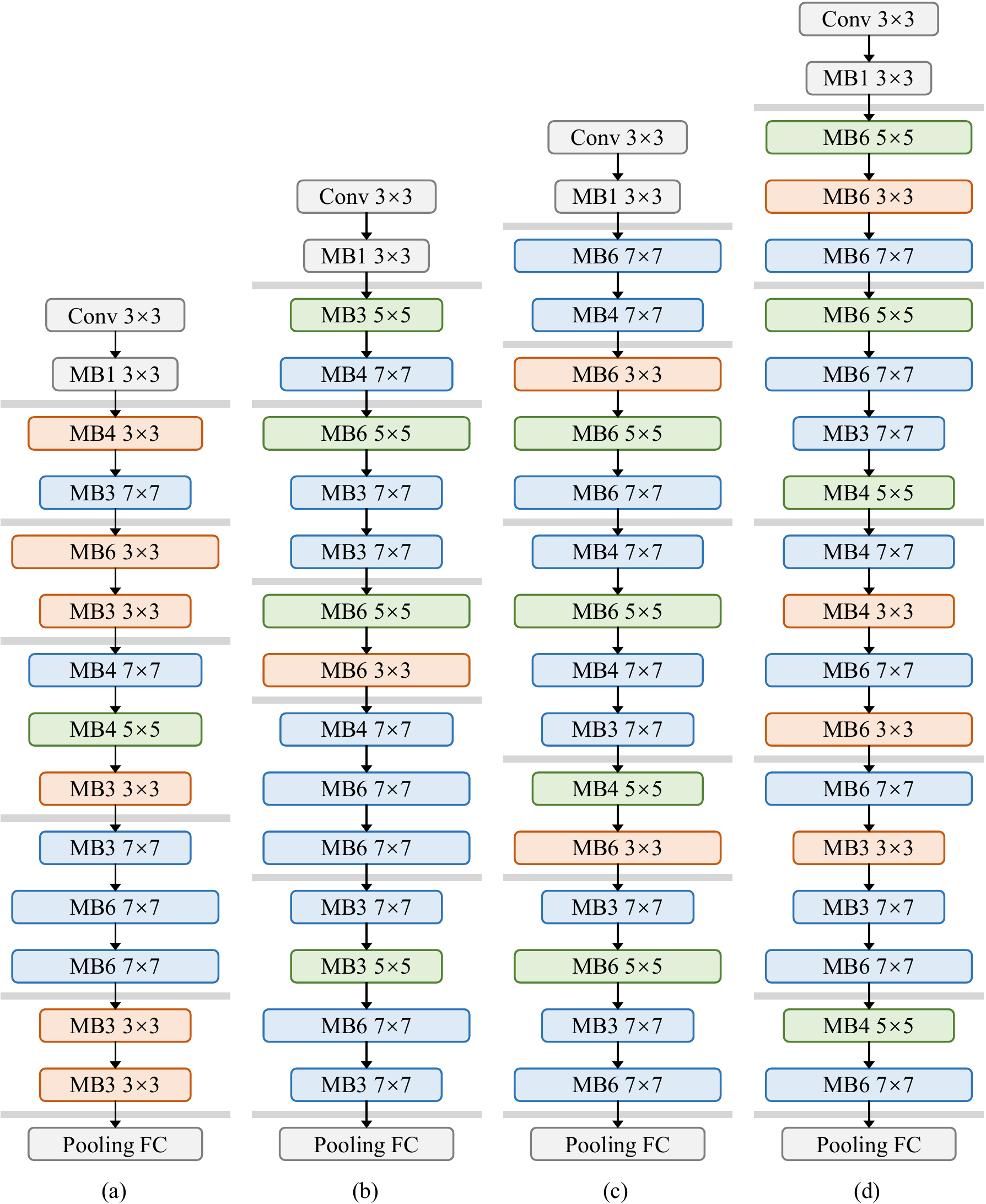}
    \caption{The architectures searched by \methodshortname for T-4 under various resource budgets on ImageNet-10. (a): The searched architecture at the \#MAdds level of 219 M. (b): The searched architecture at the \#MAdds level of 269 M. (c): The searched architecture at the \#MAdds level of 316 M. (d): The searched architecture at the \#MAdds level of 361 M.
    }
    \label{fig:task3_flops}
\end{figure}

\begin{figure}[!t]
    \centering
    \includegraphics[width=0.95\linewidth]{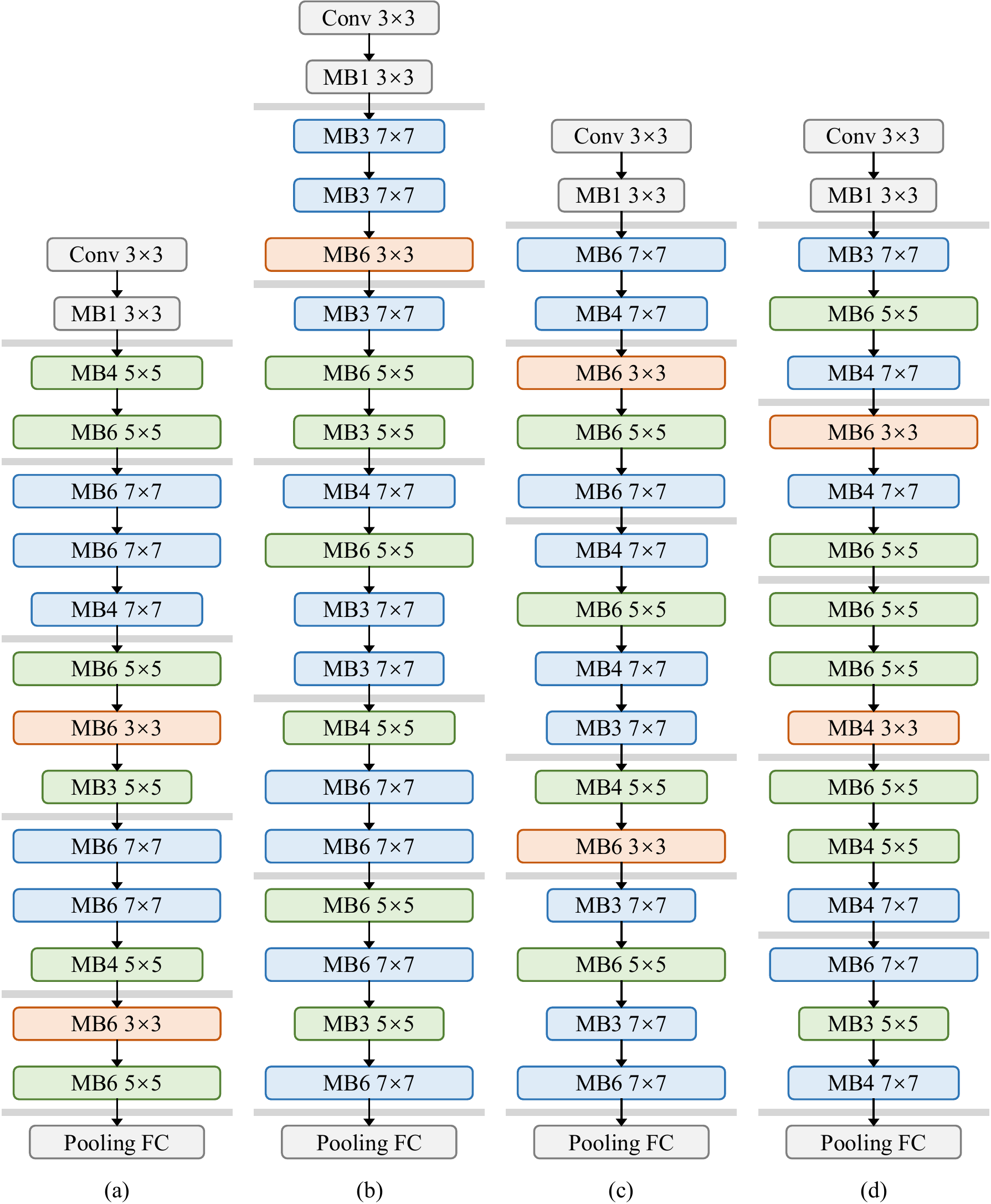}
    \caption{The architectures searched by \methodshortname at the \#MAdds level of 316 M for \qdmy{different} superclasses on ImageNet-10. (a): The searched architecture for T-1. (b) The searched architecture for T-3. (c): The searched architecture for T-4. (d): The searched architecture for T-7.
    }
    \label{fig:flops_tasks}
\end{figure}

To demonstrate the effectiveness of the proposed method,
we first compute the cosine similarity for each pair of the architectures among different superclasses under the same \#MAdds level on ImageNet-10. Here, we use the one-hot encoding to encode the architecture following~\cite{Cai2020Once_for_All}.
We then compute the average cosine similarity for each \#MAdds level and report the results in Figure~\ref{fig:simi_arch}. From the results, the average cosine similarity first increases and then saturates with the increase of \#MAdds. 
One possible reason is that the searched network tends to use larger kernel sizes, higher width expansion ratios, and more layers with the increase of \#MAdds, which is shown in Figure~\ref{fig:task3_flops}. Therefore, the searched architectures for different superclasses become \qdmy{more} similar with the increase of \#MAdds. We further show the searched architectures for different superclasses under the same level of \#MAdds in Figure~\ref{fig:flops_tasks}. From the results, the searched architectures for different superclasses differ a lot, which demonstrates the motivation that the optimal architectures for diverse superclasses are different.

\end{document}